\documentclass[conference]{IEEEtran}
\IEEEoverridecommandlockouts
\usepackage{cite}
\usepackage{amsmath,amssymb,amsfonts}
\usepackage{algorithmic}
\usepackage{graphicx}
\usepackage{textcomp}
\usepackage{xcolor}
\usepackage{booktabs,nicematrix,makecell}
\usepackage{xparse}
\usepackage{multirow}
\usepackage{lscape}
\usepackage{graphicx}
\usepackage{pifont}
\usepackage{tcolorbox}
\usepackage{booktabs}
\usepackage{hyperref}
\usepackage{enumitem}
\usepackage{tabularx}
\usepackage{colortbl}
\usepackage{array}
\usepackage{wasysym}
\usepackage{cleveref}
\usepackage{acronym}
\usepackage{amssymb}
\usepackage{threeparttable}
\usepackage{float}
\usepackage{wrapfig}
\usepackage{rotating}
\usepackage{cleveref}
\usepackage{balance}
\usepackage{dcolumn}
\newcolumntype{P}[1]{D{$\pm$}{$\pm$}{#1}}

\usepackage{siunitx}
\usepackage{etoolbox}
\usepackage{booktabs}
\usepackage{graphicx}
\robustify\bfseries

\newcommand{\para}[1]{\vskip 4pt\noindent\textbf{#1}\hskip .05in}

\newif\ifshowcomments
\showcommentsfalse
\showcommentstrue 

\def\BibTeX{{\rm B\kern-.05em{\sc i\kern-.025em b}\kern-.08em
    T\kern-.1667em\lower.7ex\hbox{E}\kern-.125emX}}
\begin{document}

\title{Smudged Fingerprints: A Systematic Evaluation of the Robustness of AI Image Fingerprints\thanks{This work has been accepted for publication in the 4th IEEE Conference on Secure and Trustworthy Machine Learning (IEEE SaTML’26). The final version will be available on IEEE Xplore.}}



\author{
\IEEEauthorblockN{Kai Yao}
\IEEEauthorblockA{\textit{School of Informatics}\\
\textit{University of Edinburgh}\\
\href{mailto:kai.yao@ed.ac.uk}{kai.yao@ed.ac.uk} \\}
\and
\IEEEauthorblockN{Marc Juarez}
\IEEEauthorblockA{\textit{School of Informatics}\\
\textit{University of Edinburgh}\\
\href{mailto:marc.juarez@ed.ac.uk}{marc.juarez@ed.ac.uk} \\}
}

\maketitle

\begin{abstract}
Model fingerprint detection has shown promise to trace the provenance of AI-generated images in forensic applications. However, despite the inherent adversarial nature of these applications, existing evaluations rarely consider adversarial settings. We present the first systematic security evaluation of these techniques, formalizing threat models that encompass both white- and black-box access and two attack goals: fingerprint removal, which erases identifying traces to evade attribution, and fingerprint forgery, which seeks to cause misattribution to a target model. We implement five attack strategies and evaluate 14 representative fingerprinting methods across RGB, frequency, and learned-feature domains on 12 state-of-the-art image generators. Our experiments reveal a pronounced gap between clean and adversarial performance. Removal attacks are highly effective, often achieving success rates above $80\%$ in white-box settings and over $50\%$ under black-box access. While forgery is more challenging than removal, its success varies significantly  across targeted models. We also observe a utility--robustness trade-off: accurate attribution methods are often vulnerable to attacks and, although some techniques are robust in specific settings, none achieves robustness and accuracy across all evaluated threat models. These findings highlight the need for techniques that balance robustness and accuracy, and we identify the most promising approaches toward this goal. Code available at: \texttt{\url{https://github.com/kaikaiyao/SmudgedFingerprints}}.

\end{abstract}


\section{Introduction}

The widespread adoption of generative AI models for image generation has created an urgent need for content provenance mechanisms.
These models---now integrated into creative tools~\cite{adobe2023firefly},
content platforms~\cite{katsamakas2024gcp}, and commercial APIs~\cite{openai2023dalle3}---can produce images that are almost visually indistinguishable from authentic photographs. While these technological advances enable new forms of creativity, they also come with risks for misinformation, fraud, and abuse, particularly when AI-generated content is passed as genuine~\cite{bird2023typology}. As a consequence, reliably identifying and tracing generated images has become a pressing challenge for maintaining trust and accountability in digital media.

A promising avenue to address this challenge builds on a key property of these models: they tend to leave artifacts---subtle but consistent patterns---in the images they produce~\cite{Marra18}. Initially, these patterns were studied as indicators that the images were AI-generated, but researchers showed that the patterns vary across different models and may even serve as fingerprints of the models~\cite{Marra19a}. These discoveries have led to the development of \emph{model fingerprint detection} (MFD), a set of techniques that exploit these fingerprints to address two related problems: 
\emph{deepfake detection}---determining whether an image is AI-generated---and \emph{model attribution}---identifying which specific model generated it.

Existing MFD techniques evaluated under various settings achieve over 90\% accuracy for both deepfake detection~\cite{Marra19a,Nataraj19,Durall20,Dzanic20,Wang20,Corvi23} and model attribution~\cite{Marra18,Marra19b,Yu19,Wang20,Guarnera20,Girish21}.
If these accuracies were to hold in realistic scenarios, MFD would prove useful in bridging the accountability gap in current deployments of generative models. Indeed, being able to identify deepfakes would help mitigate the risks of misinformation and scams, while attributing a harmful output to the specific model that generated it would make it possible to flag issues, understand their causes, and prevent similar issues in the future.

However, high detection accuracy alone is not sufficient to guarantee the practical utility of these techniques. If MFD is to be used as a tool to detect model misuse and be a deterrent for malicious actors, as the academic literature suggests~\cite{Marra19a,Marra19b}, these techniques must also be reliable in adversarial settings. For example, to prevent the destruction of forensic evidence, it is critical that fingerprints cannot be easily removed.

Yet, security evaluations of MFD are scarce and limited. The few existing studies only evaluate a small subset of MFD techniques, consider narrow threat models and attacker goals, or evaluate only simple image transformations, failing to consider strategic adversaries who are aware of existing MFD techniques and actively attempt to circumvent them~\cite{goebel2020adversarial,Wesselkamp22}.
This lack of security consideration is in stark contrast to alternative techniques that have been designed for the same purpose. Model watermarking methods, for example, take a more active approach by embedding detectable signals in the generated images, rather than relying on naturally occurring artifacts like MFD~\cite{boenisch2021systematic}. While watermarking schemes have been developed with a strong emphasis on security, including systematic evaluations of their robustness to removal and forgery attacks~\cite{lin2025crack,Jiang2023EvadingWatermark,Hu2024StableSignature,Zhao2024InvisibleRemovable, an2024waves}, the robustness of MFD to similar attacks has received little attention in comparison.

In this paper, we address this gap by providing the first systematic security evaluation of MFD techniques for model attribution. Our evaluation framework includes three plausible threat models that capture varying levels of adversarial knowledge. These threat models lead to five attack strategies that we implement and evaluate on a comprehensive set of MFD techniques compiled from the literature, spanning different feature extraction approaches (pixel-domain, frequency-domain, learned representations), detection algorithms (statistical methods, neural networks), and evaluation settings.

We focus on model attribution rather than deepfake detection because model attribution is the harder task for MFD: it requires not merely detecting the presence of model-induced patterns, but distinguishing architecture-specific differences across them, which requires higher entropy.
Therefore, evaluating MFD techniques for model attribution allows us to draw conclusions for deepfake detection: if the techniques prove robust against attacks on model attribution, they should remain robust in detecting  for the simpler task of deepfake detection.

Our threat models consider two attacker goals: fingerprint removal and fingerprint forgery---the two attacks that have been evaluated in the literature on watermarking for deepfake detection.
Within the context of model attribution, removal refers to the suppression of the fingerprint signal such that an image is no longer attributed to its true source, enabling malicious actors to disguise images generated through banned or problematic models~\cite{bbc_deepnude_2019}; fingerprint forgery aims to alter the fingerprints such that the images are misattributed to different models, potentially allowing adversaries to falsely implicate legitimate providers.

Our findings show that existing fingerprinting techniques are highly vulnerable to removal attacks, even under constrained attacker capabilities, while success of forgery attacks is more limited. The effectiveness of removal attacks is concerning by itself as it severely limits the utility of these techniques for the applications that have been advocated in the literature~\cite{Marra19a,Marra19b}. These results highlight the need for more robust MFD techniques and alternative accountability mechanisms to ensure reliable model attribution in adversarial settings.

\medskip\noindent To summarize, our main contributions are:

\begin{itemize}
    \item We conduct a comprehensive review of the literature on MFD methods across three domains: RGB, frequency, and learned representations, and identify gaps in existing security evaluations.  
    \item We formalize threat models based on three levels of adversarial knowledge for both removal and forgery attacks, and derive five concrete attack strategies for implementation.  
    \item We design and implement an evaluation framework that executes these attacks, benchmarking 14 representative MFD methods from the literature for a model attribution task across 12 state-of-the-art generative models spanning GANs, VAEs, and diffusion models on FFHQ dataset.  
    \item We provide an in-depth analysis and discussion of the attack results, and identify promising directions for strengthening the robustness of MFD methods.  
\end{itemize}

\section{Background and Related Work}

We begin by reviewing existing model fingerprint detection (MFD) techniques for image generation models, along with evaluations of their security properties. This overview highlights key research gaps, particularly regarding the robustness of these techniques for model attribution, gaps that our evaluation framework is designed to address.

\subsection{Model Fingerprint Detection}
Model fingerprint detection (MFD) has gained significant attention as a passive approach to identifying images produced by generative models, such as GANs, VAEs, and diffusion models. Unlike watermarking, which embeds signals during generation, MFD relies on detecting inherent artifacts or ``fingerprints'' left by the model's processes. These fingerprints can be detected in statistical irregularities of color and frequency patterns in the image.

Following the classification in Song et al.~\cite{Song24}, we group MFD methods into those extracting features from the RGB pixel domain, frequency domain, learned features through deep learning, and cross-domain approaches. Across these categories, methods have shown strong performance in controlled settings, often achieving near-perfect accuracy on specific datasets, but their generalization to unseen models or post-processed images varies.

Most of the literature focuses on MFD for deepfake detection, but several studies also evaluate the use of MFD fingerprints for model attribution.
While deepfake detection is a binary task, model attribution requires discerning subtle, model-specific patterns among similar architectures, making it a more challenging task. Yet, most evaluations of MFD for model attribution report high detection performance.

\para{RGB Pixel Domain.}
Methods in this category analyze spatial patterns and color statistics in the image's pixel values, exploiting deviations from natural image distributions introduced by the generative processes. For instance, early work focused on color saturation irregularities, where the use of normalization layers in generative models tends to suppress over- or under-exposed pixels. One of the earliest MFD approaches applied this strategy for deepfake detection, reporting an AUC of 0.70 for fully GAN-generated faces and 0.61 for partially GAN-synthesized faces, evaluated on a benchmark of roughly 10,000 images from models such as CycleGAN~\cite{McCloskey18}. Building on this idea, Nataraj et al.\ use per-channel co-occurrence matrices to capture inter-channel pixel dependencies, achieving over 99\% accuracy on more than 56,000 CycleGAN and StarGAN images~\cite{Nataraj19}.
Later work extended this to cross-channel co-occurrence, improving deepfake detection of high-quality GAN faces on datasets like CelebA~\cite{Nowroozi22}. Although effective for deepfake detection, these methods cannot be directly applied to model attribution without modifications.

\para{Frequency Domain.}
Generative models often introduce spectral anomalies, such as deficiencies in high-frequency components or unnatural power distributions, which frequency-based methods target through transformations like FFT or DCT. For example, approaches using azimuthally averaged power spectral densities from 2D FFTs have demonstrated near-perfect discrimination (close to 100\% accuracy) for deepfake detection between GAN outputs and real images on CelebA and LSUN, detecting patterns in high-frequency components for models like DCGAN and CycleGAN~\cite{Durall20}.
Other methods extract features from Fourier modes, such as decay exponents and magnitudes, achieving up to 99\% accuracy for deepfake detection on ProGAN-generated faces~\cite{Dzanic20}.
Giudice et al.\ instead rely on DCT analysis, extracting a 63-dimensional fingerprint vector from the means of AC coefficients of image blocks, achieving an AUC above 0.95 for deepfake detection on datasets like FaceForensics++~\cite{Giudice21}. More recent work combines noise residual autocorrelations with radial and angular spectrum averaging, achieving over 90\% model attribution accuracy across GANs, VAEs, and diffusion models on diverse benchmarks including FFHQ and ImageNet subsets~\cite{Corvi23}.

\para{Learned Features Domain.}
Deep learning approaches automatically extract implicit fingerprints from high-dimensional representations, and have been widely explored for both deepfake detection and model attribution. For deepfake detection, Marra et al.~\cite{Marra18} trained CNNs such as InceptionNet~v3 and XceptionNet on social media-like images, achieving around 95\% accuracy in distinguishing GAN-generated images from real ones across datasets containing CycleGAN and StarGAN outputs. In the model attribution setting, Wang et al.\ demonstrated that with careful pre- and post-processing and extensive data augmentation, a ResNet-50 trained on a single generator (ProGAN) can generalize effectively to 11 unseen GAN architectures, reaching 99\% detection accuracy~\cite{Wang20}.
Complementary work has extended CNNs with frequency-domain branches, achieving AUC $>$ 0.98 on FaceForensics++ and CelebA-DF~\cite{Frank20,Qian20}. Yu et al.~\cite{Yu19} proposed CNN-based methods that learn model-specific fingerprints, reporting up to 98\% attribution accuracy on CelebA across models including ProGAN and StyleGAN, while Girish et al.~\cite{Girish21} provided one of the few open-world evaluations, achieving 85--90\% recall over 20 GAN architectures---including previously unseen models---using an adaptive clustering algorithm. To improve adaptability, Marra et al.\ proposed incremental learning strategies that enable detectors to handle new architectures without requiring full retraining, achieving AUC $>$ 0.9 across up to five GAN models~\cite{Marra19b}. Taken together, these results highlight that deep feature-based approaches remain among the most effective techniques for both deepfake detection and model attribution.

\para{Cross-Domain Approaches.}
To improve performance, Song et al.\ introduced ManiFPT, a cross-domain fingerprinting framework that unifies multiple domains—RGB, frequency, and learned features~\cite{Song24}.
The framework defines model fingerprints as deviations of generated images from the true data manifold, formalizing the intuition that the artifacts these models leave in the images are the result of imperfections in modeling the real data distribution.
This approach enabled improved model attribution across GANs, VAEs, and diffusion models on benchmarks such as CIFAR-10 and CelebA. A subsequent extension incorporated Riemannian geometry to leverage non-Euclidean manifolds, leading to an average gain of about 6\% in model attribution accuracy~\cite{Song25}.

\subsection{Existing Robustness Evaluations}
While MFD methods perform well in benign settings, their robustness against adversarial evasion remains underexplored—particularly for model attribution, where attackers may attempt to remove or forge subtle, model-specific fingerprints. Existing evaluations mostly test robustness against generic image perturbations or targeted fingerprint removal, typically in black-box settings that do not adapt to the detector’s architecture. Moreover, most studies emphasize deepfake detection rather than model attribution. Regarding the study of attacks, prior work has almost exclusively focused on fingerprint removal, overlooking forgery attacks in which adversaries mimic legitimate fingerprints to mislead attribution.

\para{Generic Image Perturbations.}
Most evaluations evalate MFD robustness to generic image perturbations such as JPEG compression, blurring, resizing, and rotation, which can unintentionally degrade fingerprints. In the spatial pixel domain, Fourier spectrum–based methods have shown that JPEG compression can reduce detection accuracy by 10–20\% on ProGAN datasets, primarily due to the loss of high-frequency components~\cite{Dzanic20}. DCT-based approaches demonstrate partial robustness, with accuracy declining by only 5–10\% under scaling and rotation on FaceForensics++ (original AUC $>$ 0.95)~\cite{Giudice21}. CNN-based detectors are more resilient under moderate JPEG compression, retaining around 90\% accuracy on social media-style datasets with CycleGAN images~\cite{Marra18}. Spectrum-domain perturbations, such as high-pass filters, have also been evaluated; for example, detectors like Frank et al.'s~\cite{Frank20} experienced up to a 30\% accuracy drop on CelebA when subjected to such filters~\cite{Wesselkamp22}. Overall, black-box evaluations suggest that although some methods tolerate mild perturbations, others exhibit significant performance degradation even without adaptive adversarial strategies.

\para{Method-Specific Attacks.}
A few studies have investigated adaptive attacks specifically designed against MFD techniques, aiming to suppress fingerprints while maintaining perceptual quality. For co-occurrence-based detectors such as Nataraj et al.'s~\cite{Nataraj19}, projected gradient descent (PGD) approximations of the co-occurrence operator can adjust pixel statistics to resemble those of real images, reducing detection accuracy from over 99\% to nearly 0\% in white-box settings on CycleGAN datasets~\cite{Goebel20}. Similarly, frequency-targeted attacks, including peak suppression and mean-spectrum subtraction, have been shown to evade detectors like Frank et al.'s~\cite{Frank20}, lowering accuracy by 50–80\% for deepfake detection on ProGAN and StyleGAN images~\cite{Wesselkamp22}. These results demonstrate the vulnerability of MFD methods to adaptive adversaries. However, the extent of their susceptibility to adaptive attacks, particularly against forgery and within model attribution settings, remains largely underexplored.

\subsection{Research Gaps} 

Most MFD studies overlook the security of the techniques despite that the authors propose inherently adversarial applications. For example, uses of fingerprinting in forensics must consider attackers who aim to erase the fingerprints.

We identify four major limitations in the few existing security evaluations:

\begin{enumerate}
    \item These studies evaluate only simple image perturbations, overlooking strategic adversaries with knowledge of the MFD technique. Since adaptive attacks are more likely to succeed, these evaluations give a false sense of security.

    \item The only two evaluations that include adaptive attacks evaluate the attacks against specific MFD techniques, instead of using appropriate baselines or evaluating against a comprehensive range of techniques. An evaluation on a single technique does not allow to extrapolate to the effectiveness of the attacks against other techniques.

    \item Despite the fact that these studies propose MFD techniques for model attribution, most evaluations only assess security for deepfake detection, leaving security regarding model attribution unaddressed.

    \item  All prior research has focused on removal attacks, leaving robustness to forgery attacks underexplored. As we explain in Section~\ref{sec:adv_obj}, although these two attacks are related, success of one does not imply the other.
\end{enumerate}

Our work closes this gap by formalizing threat models for MFD techniques and systematically deriving attack strategies from these threat models. To provide comprehensiveness, we group MFD techniques with similar characteristics and evaluate them under a common evaluation framework that implements all our variations of removal and forgery attacks.


\section{Threat Model}
\label{sec:threat_model}

To systematically evaluate the security of model fingerprint detection (MFD) techniques for model attribution tasks, we formalize adversary models with varying levels of knowledge and access—ranging from a full black-box setting, where the adversary has no information about the MFD system, to a white-box setting, where the adversary has complete access. Under these different visibility levels, we study two objectives: \emph{fingerprint removal} and \emph{fingerprint forgery}. Each access level motivates distinct attack strategies, which we implement and evaluate in later sections.

\subsection{System Model}
Let $G : \mathcal{Z} \rightarrow \mathcal{X}$ denote a generative model that maps latent variables $z \in \mathcal{Z}$ to images $x = G(z) \in \mathcal{X}$. We define a \emph{model fingerprint detection (MFD) system for model attribution} as
\begin{equation}
    \mathcal{F}(x) = h(\phi(x)),
\end{equation}
where $\phi : \mathcal{X} \rightarrow \mathcal{A}$ is the \emph{fingerprint extractor} that derives fingerprint features from the input image, and $h : \mathcal{A} \rightarrow \mathcal{Y}$ is the \emph{model attribution classifier} that predicts the source identity $y \in \mathcal{Y} = \{G_1, G_2, \ldots, G_K\}$ among $K$ candidate generators.  

This formulation---decoupling the fingerprint extractor $\phi$ from the attribution classifier $h$---matches the structure of most MFD methods in the literature. Some approaches instead employ end-to-end neural networks, where fingerprint extraction is not a separate function (e.g.,~\cite{Qian20,Wang20}). In such cases, the fingerprint can be defined implicitly: the neural network except the last layer acts as the fingerprint extractor $\phi$, and the last layer can be interpreted as the model attribution classifier $h$.

\subsection{Adversarial Objectives}\label{sec:adv_obj}
The adversary aims to generate an adversarial image $x'$ that misleads the MFD system $\mathcal{F}$. We impose a general \emph{distance constraint} between the original and adversarial images:
\begin{equation}\label{eq:dist_constraint}
    d(x, x') \leq \varepsilon,
\end{equation}
where $d:\mathcal{X}\times\mathcal{X}\to\mathbb{R}_{\geq 0}$ is a distance function and $\varepsilon$ specifies the perturbation budget. Typical choices of $d$ include $\ell_\infty$ and $\ell_2$ norms, LPIPS~\cite{zhang2018unreasonable}, and PSNR. This constraint ensures that adversarial manipulations do not degrade the quality of the attacked images excessively.

We consider two adversarial objectives under this constraint:

\begin{itemize}
    \item \textbf{Fingerprint Removal.}  
    Given an image $x = G(z)$ with true source label $y = G$, the attacker creates $x'$ such that
    \begin{equation}\label{eq:removal}
        \mathcal{F}(x') \neq y, \quad d(x, x') \leq \varepsilon.
    \end{equation}
    A successful removal attack suppresses enough model-specific information in the fingerprint features to cause the model attribution classifier $h$ to misclassify $x'$. Since the MFD system operates in a closed world of candidate generators, such misclassification typically results in a low-confidence attribution to a different model.
    
    \item \textbf{Fingerprint Forgery.}  
    Given $x = G(z)$ with true source label $y = G$ and a chosen target label $y_t = G_t \neq y$, the adversary produces $x'$ such that
    \begin{equation}\label{eq:forgery}
        \mathcal{F}(x') = y_t, \quad d(x, x') \leq \varepsilon.
    \end{equation}
    Unlike removal, forgery attacks are \emph{targeted}: the adversary induces a specific misclassification, forcing the MFD system to attribute $x'$ to the target generator $G_t$ rather than the true source $G$. This requires introducing sufficient features resembling $G_t$’s fingerprints so that the model attribution classifier $h$ outputs the target label $y_t$.
\end{itemize}

\subsection{Adversary Knowledge Levels}
We distinguish three levels of adversarial knowledge about the MFD system $\mathcal{F}$, depending on the adversary’s access to the fingerprint extractor $\phi$, the model attribution classifier $h$, and their knowledge of the model attribution task:

\begin{enumerate}
    \item \textbf{White-box Access.}  
    The adversary has complete access to both $\phi$ and $h$, enabling end-to-end gradient computations through $\mathcal{F}$. This represents the strongest adversary and, as our results show, allows for highly effective removal and forgery attacks. While less common, this setting is realistic when the MFD system is open-sourced or if a malicious insider gains direct access to the target model.
    
    \item \textbf{Black-box Access I.}  
    The adversary has no access to $\phi$ or $h$, but knows the model attribution task, i.e., the candidate generators $\mathcal{Y}$. By sampling images from all $G \in \mathcal{Y}$, the adversary can train a surrogate classifier $h_s$ directly on generated images, bypassing fingerprint extraction. Adversarial examples $x'$ are then crafted against $h_s$ with the expectation that they transfer to the true MFD system $\mathcal{F}$, leveraging adversarial transferability~\cite{goodfellow2014explaining}. This scenario is realistic since attribution systems are typically concealed to prevent manipulation. Attackers can only use publicly available generators to build surrogates, not to replicate the defender's pipeline but to create adversarial examples that may transfer to the MFD system.
    
    \item \textbf{Black-box Access II.}  
    The adversary has no knowledge of $\phi$, $h$, or the model attribution task. Typical low-cost attacks in this setting are generic image manipulations such as compression, noise, or blurring. This setting makes the least assumptions and is also realistic: defenders usually hide system details, and opportunistic attackers may only apply low-cost transformations in the hope of reducing model attribution confidence.
\end{enumerate}

Across all three knowledge levels, we assume the adversary does not have internal access to the target generator $G$ (nor to $G_t$ in forgery), but can query any of the candidate generators and sample images from them. This reflects real-world deployments of generative models that expose subscription-based access through user APIs.
Internal access to the generator, however, represents a stronger threat model outside our scope: although an attacker compromising model provider infrastructure may obtain internal information about the generator, such access enables threats far more severe than fingerprint forgery or removal attacks, making these attacks moot.
In addition, although access to a generator might give the adversary an advantage against an MFD method designed for that specific generator, most MFD methods---like our attacks---are generator-agnostic and applicable across generator families.

Removal attacks are feasible under all three knowledge levels, while forgery attacks are infeasible in the Black-box II setting, as the adversary does not have access to $G_t$, and thus lacks the information needed to cause a targeted misattribution.

\section{Attack Implementation}
\label{sec:attack_impl}

Having formalized the threat models, we now describe the concrete implementations of attack strategies against an MFD system for model attribution.
We propose five attack strategies, each motivated by the adversary's knowledge level and the mathematical properties of the fingerprint extractor $\phi$. These are general strategies, not tied to any single MFD implementation; they apply whenever the stated assumptions about $\phi$ and the adversary's access hold. This makes our evaluation framework broadly applicable to unseen MFD methods.

\subsection{White-box Attack Strategies}
Under \textbf{White-box Access}, the adversary has full access to both $\phi$ and $h$. Because $h$ is typically a differentiable neural-network classifier, and $\mathcal{F}$ becomes differentiable whenever $\phi$ is differentiable (or can be made so), the adversary can apply \emph{gradient-based optimization} to craft adversarial examples~\cite{goodfellow2014explaining,madry2017towards}. Concretely, the attacker computes gradients of an attribution loss of $\mathcal{F}$ w.r.t.\ the input image and iteratively perturbs pixels to reduce the confidence of the correct class (removal) or to increase the confidence of a chosen target class (forgery).

Depending on properties of $\phi$, we distinguish three white-box optimization strategies:

\begin{itemize}
    \item \textbf{W1: Direct Differentiation.}  
    When $\phi$ is differentiable, attackers directly apply gradient-based attacks on $h(\phi(x))$.
    \item \textbf{W2: Analytic Approximation ($\tilde{\phi}$).}  
    If $\phi$ is non-differentiable but admits a differentiable analytic approximation, the adversary constructs a differentiable surrogate $\tilde{\phi}(x)$ (e.g., soft histograms~\cite{Yusuf2020DifferentiableHistogram}, relaxed co-occurrence statistics~\cite{Levy2014NeuralWordEmbedding}) and optimizes $h(\tilde{\phi}(x))$.
    
    \item \textbf{W3: Surrogate Extractor ($\phi_s$).}  
    If $\phi$ is neither differentiable nor amenable to analytic approximation, the adversary trains a differentiable surrogate extractor $\phi_s(x)$ using pairs $\{(x, \phi(x))\}$ sampled from generators $G \in \mathcal{Y}$ to approximate $\phi$. Attacks then optimize $h(\phi_s(x))$.
\end{itemize}

\begin{figure}[!htbp]
    \centering
    \includegraphics[width=0.8\columnwidth]{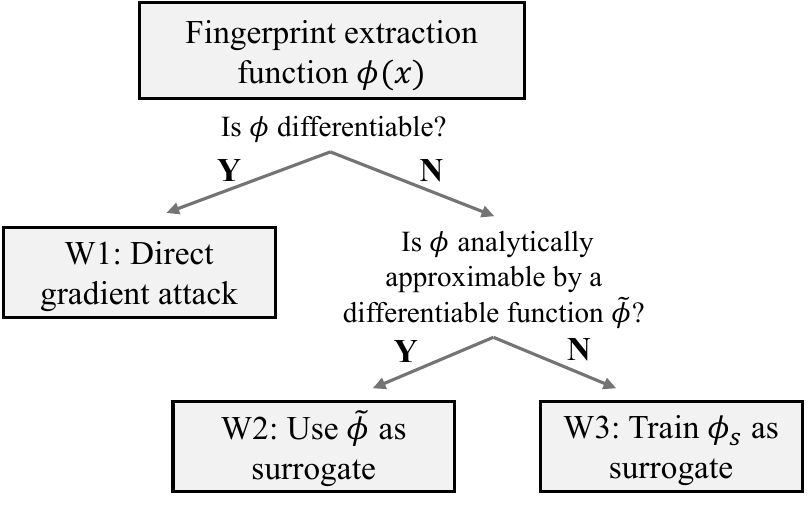}
    \caption{Decision tree for selecting a white-box attack strategy. The applicability of white-box attacks depends on whether the fingerprint extractor, $\phi$, is differentiable and, if not, whether it can be analytically approximated.}
    \label{fig:attack_types_logic}
\end{figure}

Figure~\ref{fig:attack_types_logic} depicts which white-box attacks (W1--W3) apply given properties of $\phi$. These strategies differ in their assumptions: W1 is the strictest, requiring perfect gradient information. W2 and W3 are possible when there is gradient information but offer no advantage over W1. Similarly, W3 is possible when W2 is feasible, but not the other way around.

\subsection{Black-box Attack Strategies}
When the attacker has limited knowledge of $\phi$ and $h$, attacks rely on surrogate training or generic image manipulations:

\begin{itemize}
    \item \textbf{B1: Surrogate Attribution Classifier ($h_s$).}  
    Under \textbf{Black-box Access I}, the candidate set $\mathcal{Y}$ is known but $\phi$ is not. The adversary can sample images from each $G \in \mathcal{Y}$ and train a surrogate classifier $h_s(x)$ that approximates the mapping $x \mapsto h(\phi(x))$ on the image space. The adversary then applies gradient-based attacks on $h_s$, hoping that they transfer to the targeted MFD system. Note that the surrogate $\phi_s$ in W3 is a surrogate of the fingerprint extractor $\phi$, while this surrogate, $h_s$, is for the entire model attribution classifier $h$.
    
    \item \textbf{B2: Image Transformations.}  
    Under \textbf{Black-box Access II}, without knowledge of $\phi$, $h$, or $\mathcal Y$, a typical adversary applies generic image transformations such as JPEG compression, noising, blurring, or resizing. These untargeted manipulations aim to degrade fragile fingerprint traces and reduce attribution confidence.
\end{itemize}

Under Black-box Access II, adversaries could pursue more sophisticated attacks, such as training a surrogate on outputs from the target generator and a dataset for the negative class.
However, these require substantial sampling and optimization, and would still underperform B1 under Black-box Access I (with knowledge of $\mathcal Y$). For this reason, we restrict Black-box Access II to low-cost image perturbations as a representative, practical attack strategy.


\subsection{Unified Optimization Objective}
All gradient-based attacks (W1–W3, B1) can be expressed within a single constrained optimization framework:
\begin{equation}
    \min_{x'} \;\; \mathcal{L}_{\text{adv}}(\mathcal{F}'(x')) 
    \quad \text{subject to} \quad d(x, x') \leq \varepsilon,
    \label{eq:general_attack}
\end{equation}
where the constraint is the same as in Eq.~\ref{eq:dist_constraint}. Here, $\mathcal{F}'$ denotes the attribution model available to the attacker in each setting:
\begin{itemize}
    \item W1: $\mathcal{F}'(x) = \mathcal{F} = h(\phi(x))$, i.e., the true MFD system.  
    \item W2: $\mathcal{F}'(x) = h(\tilde{\phi}(x))$, where $\tilde{\phi}$ is an analytic differentiable approximation of $\phi$.  
    \item W3: $\mathcal{F}'(x) = h(\phi_s(x))$, where $\phi_s$ is a learned surrogate extractor approximating $\phi$.  
    \item B1: $\mathcal{F}'(x) = h_s(x)$, where $h_s$ is a surrogate attribution classifier trained on images.  
\end{itemize}
The definition of the adversarial loss $\mathcal{L}_{\text{adv}}$ depends on the attack objective:

\begin{itemize}
    \item \textbf{Fingerprint Removal:}  
    suppress attribution to the true source $y$ by optimizing a loss that discourages high confidence on that label.
    We use a cross-entropy–based objective:
    \[
        \mathcal{L}_{\text{adv}} = - \log \left( 1 - P_{\mathcal{F}'}(y \mid x') \right),
    \]
    which provides stable gradients and effectively drives the prediction away from $y$.

    \item \textbf{Fingerprint Forgery:}  
    induce attribution to a chosen target generator with label $y_t$.  
    We minimize the cross-entropy loss with respect to the target:
    \[
        \mathcal{L}_{\text{adv}} = - \log P_{\mathcal{F}'}(y_t \mid x'),
    \]
    encouraging confident misattribution toward $y_t$ and yielding more stable and effective optimization.
\end{itemize}

\para{Summary.} The attack implementation taxonomy detailed in this section provides an extendable framework for evaluating and comparing removal and forgery robustness across MFD techniques and adversary knowledge levels. We will introduce our experimental setup in Section~\ref{sec:setup} and empirically evaluate all attack strategies in Section~\ref{sec:results}.

\begin{table*}[!htbp]
\centering
\caption{MFD techniques for generative model attribution. The table shows domain, fingerprint description, size, original attribution accuracy, mathematical properties (differentiability $\nabla_x \mathcal{\phi}$ and analytic approximation $\tilde{\phi}$), and attack compatibility (W1: direct gradient attacks, W2: analytic approximation attacks, W3: surrogate extractor attacks, B1-B2: black-box attacks).}
\label{tab:fingerprint_method_properties}
\renewcommand{\arraystretch}{1.0}
\setlength{\tabcolsep}{2.5pt}
\begin{tabular}{l l l l c c c c c c c c}
\toprule
\textbf{Method} & \textbf{Domain} & \textbf{Fingerprint Description} & \textbf{Size} & \textbf{Accuracy (\% $\pm$ Std)} & \textbf{$\nabla_x \mathcal{\phi}$} & \textbf{$\tilde{\phi}$}
& \textbf{W1} & \textbf{W2} & \textbf{W3} & \textbf{B1} & \textbf{B2} \\
\midrule
McCloskey18~\cite{McCloskey18}   & RGB        & Saturation pixel proportions & 1×4       & 23.81 $\pm$ 0.94 & $-$ & \checkmark &          & $\bullet$ & $\bullet$ & $\bullet$ & $\bullet$ \\
Nataraj19~\cite{Nataraj19}       & RGB        & Co-occurrence matrices & 3×256×256 & 90.96 $\pm$ 0.74 & $-$ & \checkmark &          & $\bullet$ & $\bullet$ & $\bullet$ & $\bullet$ \\
Nowroozi22~\cite{Nowroozi22}     & RGB        & Cross-band co-occurrence & 6×256×256 & 87.31 $\pm$ 1.15 & $-$ & \checkmark &          & $\bullet$ & $\bullet$ & $\bullet$ & $\bullet$ \\
Song24-RGB~\cite{Song24}         & RGB        & RGB manifold deviations & 3×256×256 & 63.82 $\pm$ 0.76 & $-$ & $-$ &          &  & $\bullet$           & $\bullet$ & $\bullet$ \\
\cmidrule{1-12}
Marra19a~\cite{Marra19a}         & Frequency  & Denoising residuals & 3×256×256 & 66.67 $\pm$ 2.16 & $-$ & \checkmark &          & $\bullet$ & $\bullet$ & $\bullet$ & $\bullet$ \\
Durall20~\cite{Durall20}         & Frequency  & Radial power spectrum & 1×179     & 80.90 $\pm$ 1.02 & $-$ & \checkmark &          & $\bullet$ & $\bullet$ & $\bullet$ & $\bullet$ \\
Dzanic20~\cite{Dzanic20}         & Frequency  & Power law decay params & 1×3       & 78.98 $\pm$ 0.11 & $-$ & \checkmark &          & $\bullet$ & $\bullet$ & $\bullet$ & $\bullet$ \\
Giudice21~\cite{Giudice21}       & Frequency  & DCT coefficient statistics & 1×63      & 94.88 $\pm$ 1.04 & \checkmark & $-$ & $\bullet$  &            &            & $\bullet$ & $\bullet$ \\
Corvi23-R~\cite{Corvi23}         & Frequency  & Noise residual autocorr & 1×4225    & 39.65 $\pm$ 1.19 & \checkmark & $-$ & $\bullet$  &            &            & $\bullet$ & $\bullet$ \\
Corvi23-S~\cite{Corvi23}         & Frequency  & Radial/angular spectra & 1×144     & 27.11 $\pm$ 8.62 & \checkmark & $-$ & $\bullet$  &            &            & $\bullet$ & $\bullet$ \\
Song24-Freq~\cite{Song24}        & Frequency  & Frequency manifold deviations & 1×256×256 & 84.21 $\pm$ 0.85 & $-$ & $-$ &          &  & $\bullet$           & $\bullet$ & $\bullet$ \\
\cmidrule{1-12}
Qian20~\cite{Qian20}             & Learned         & F3-Net frequency features & $-$         & 90.12 $\pm$ 0.84 & \checkmark & $-$ & $\bullet$  &            &            & $\bullet$ & $\bullet$ \\
Wang20~\cite{Wang20}             & Learned         & ResNet50 features &  $-$        & 98.47 $\pm$ 0.51 & \checkmark & $-$ & $\bullet$  &            &            & $\bullet$ & $\bullet$ \\
Song24-SL~\cite{Song24}          & Learned         & SL manifold deviations & 1×2048    & 87.90 $\pm$ 0.91 & $-$ & $-$ &          &  & $\bullet$           & $\bullet$ & $\bullet$ \\
\bottomrule
\end{tabular}
\end{table*}

\section{Experimental Setup}
\label{sec:setup}

We now describe the experimental setup used to evaluate the attack strategies, which includes the generative models, MFD methods, and datasets selected for evaluation. We also describe the training of the attribution models and the details of our implementations of the attack strategies.  

\subsection{Image Generators}

We evaluate our framework on a model attribution task involving images from 12 widely used state-of-the-art generative models, covering 3 major architectural classes: 6 GAN models (StyleGAN2~\cite{Karras2020ADA}, StyleGAN3~\cite{Karras2021StyleGAN3}, GANformer~\cite{Hudson2021GANformer}, StyleSwin~\cite{Zhang2022StyleSwin}, R3GAN~\cite{Huang2024R3GAN}, and CIPS~\cite{Anokhin2021CIPS}), 3 VAE models (VDVAE~\cite{Child2021VDVAE}, VQVAE~\cite{van2017neural}, and NVAE~\cite{vahdat2020nvae}), and 3 diffusion models (NCSN++~\cite{Song2021ScoreSDE}, LDM~\cite{rombach2022high}, and ADM~\cite{dhariwal2021diffusion}). All models are pre-trained on the FFHQ dataset and generate 256×256 RGB images. All models are sourced from official repositories with pre-trained checkpoints to ensure reproducibility.

\subsection{MFD Methods}
We evaluate 14 MFD methods spanning the RGB, frequency, and learned feature domains. Although this set is not exhaustive, it is representative: it includes influential MFD methods that have been widely studied for model attribution (e.g., those covered in Song et al.~\cite{Song24}) and thus reflects the performance and robustness trends of the state of the art.

Table~\ref{tab:fingerprint_method_properties} summarizes their key characteristics, including the feature domain (RGB, frequency, or learned representations~\cite{Song24}), a brief description of the fingerprint features, and the fingerprint size (i.e., the output of $\phi$, or $-$ for end-to-end methods). We also report attribution accuracy of the MFD methods, obtained by reproducing their evaluations on the pool of 12 candidate generative models.

The remaining columns indicate whether the method is differentiable, whether it admits an analytic approximation, and which attack strategies are applicable. For non-differentiable methods that allow an analytic approximation, we introduce differentiable variants $\tilde{\phi}$ to enable W2 attacks.

\subsection{Models and Training Details}

We train three neural network models to evaluate fingerprint robustness across different attack settings. All models are optimized with Adam for 100 epochs on $1{,}000$ images/fingerprints per generative model, using a batch size of 32, a validation split of 20\%, and early stopping to prevent overfitting. The learning rate is fixed at $1 \times 10^{-4}$.

\smallskip
\noindent The models that require training are:
\begin{itemize}
    \item \textbf{Attribution Classifier ($h$).}  
    This model maps fingerprint features to source model labels and serves as the primary target for attacks. It is implemented as an MLP with [512, 256, 128] hidden dimensions and ReLU activations, trained with cross-entropy loss. The architecture is fixed across all MFD methods, while the input dimensionality depends on the fingerprint size of the specific method.
    
    \item \textbf{Surrogate Attribution Classifier ($h_s$).}  
    This model approximates $h$ when fingerprint features are unavailable, enabling black-box attacks (B1). It takes images as input and predicts source model labels. The architecture is a CNN with four convolutional blocks (64→128→256→512 channels), followed by fully connected layers, trained with cross-entropy loss.
    
    \item \textbf{Surrogate Fingerprint Extractor ($\phi_s$).}  
    This model approximates non-differentiable fingerprint extractors by regressing from images to fingerprint features. It uses a CNN with four convolutional blocks (64→128→256→512 channels) and fully connected layers, trained with MSE loss to match the dimensionality of the original extractor’s outputs.
\end{itemize}

\subsection{Attack Implementation and Evaluation}

We implement five adversarial attack strategies across the white-box (W1--W3) and black-box (B1--B2) threat models detailed in Section~\ref{sec:attack_impl} to evaluate fingerprint robustness. 
While removal attacks are not targeted, forgery attacks require the attacker to induce a misclassification towards a specific target candidate generator label different from the true label.
In the evaluation, we choose the target label uniformly at random on each execution of a forgery attack.

For gradient-based attacks (W1--W3, B1), we employ Projected Gradient Descent (PGD)~\cite{madry2017towards} with momentum and adaptive step size selection. We randomly sample 100 images from each generative model, and conduct attacks only on the images that are accurately classified by the true attribution model $h$. The rationale is that the attacker gains no advantage from attacking images that are already misclassified.

For each attacked image, we evaluate PGD with step sizes in $\{0.0005,0.001,0.005,0.01,0.05,0.1\}$ and consider the attack successful if any of the step sizes succeeds. The attacker can also do that in practice because they have access to images and MFD methods, so the attacker can tune the hyperparameters of their attacks for each fingerprinting technique. We then aggregate successful attempts across all images to compute the \emph{Attack Success Rate (ASR)}, which estimates the probability that the attack strategy will successfully remove or forge fingerprints in any given image.

PGD is limited to 50 iterations, optimizing the cross-entropy loss with a momentum coefficient of $0.9$, random initialization, and gradient clipping. The perturbation is bounded by $\epsilon = 0.025$ (i.e., $\ell_\infty$ norm). We adopt the $\ell_\infty$ norm for its computational efficiency and its direct interpretation as a pixel value bound, while allowing to verify that perturbations remain visually acceptable. Across all attacks, this $\epsilon$ level consistently yields LPIPS $< 0.05$ and PSNR $> 35$ dB, which indicates satisfactory perceptual quality~\cite{zhang2018lpips,huynh2008psnr} (see Appendix~\ref{apx:image_quality} for more details). We also report the relationship between $\epsilon$ and ASR and LPIPS/PSNR in Figure~\ref{fig:ablation_pgd_panel} in Appendix~\ref{apx:hyperparameters}.

All gradient-based attacks support both removal and forgery objectives. For non-gradient attacks (B2), we employ four standard image perturbations with parameter settings chosen to preserve high image quality comparable to gradient-based attacks (i.e., LPIPS $< 0.05$ and PSNR $> 35$ dB). Specifically, we apply additive Gaussian noise with $\sigma = 0.005$ (B2-Noise), Gaussian blur with $\sigma = 0.5$ using a $3{\times}3$ kernel (B2-Blur), JPEG compression at quality $95$ (B2-JPEG), and isotropic downsampling to $0.9\times$ followed by upsampling (B2-Resize). Each perturbation is applied once independently to the original image, each simulating one single removal attempt. We also conduct an ablation over these hyperparameters and report their impact on attack success in Figure~\ref{fig:ablation_panel} in Appendix~\ref{apx:hyperparameters}. Since B2's perturbations are untargeted, we evaluate them only for removal attacks.

We conduct all experiments with PyTorch on NVIDIA A100 GPUs. The reported results are the mean and standard deviation over $5$ independent runs, where each run involves training all models in the pipeline and then conducting an attack.

\section{Results}
\label{sec:results}

This section presents the results of our security evaluation. We organize it into five parts: first, we assess MFD technique robustness to removal across threat models and attack strategies; second, we repeat the same evaluation for forgery; third, we study the relationship between removal and forgery; fourth, we test whether higher utility of the MFD technique is an indicator of attack success; and, fifth, we investigate the impact of the model's type and architecture on ASR.

Overall, our evaluation reveals that existing MFD techniques are highly susceptible to adversarial manipulation. Removal attacks are particularly successful; critically, even simple B2 perturbations achieve ASR $>50\%$ for many MFD methods. The factor that dominates attack success is the MFD technique rather than the fingerprinted model, with more accurate techniques being slightly more vulnerable to the attacks.

\subsection{Robustness to Removal}

\begin{table*}[!htbp]
\centering
\caption{Attack success rates (ASR, mean~$\pm$~std, in \%) of \textbf{removal attacks} against evaluated MFD methods. White-box attacks include direct gradient-based optimization (W1), analytic approximations (W2), and surrogate fingerprint extractors (W3). Black-box attacks include surrogate classifiers (B1) and fingerprint-agnostic perturbations (B2: Gaussian noise, blur, JPEG compression, resizing). Higher ASR indicates greater vulnerability. The maximum values per method are highlighted in \textbf{bold}. The row separators indicate the domains: RGB, Frequency, and Learned.}
\label{tab:removal_attacks}
\resizebox{\textwidth}{!}{
\begin{tabular}{l *{8}{S[table-format=3.2(2.2)]}}
\toprule
\textbf{Method} & {\textbf{W1}} & {\textbf{W2}} & {\textbf{W3}} & {\textbf{B1}} & {\textbf{B2-Noise}} & {\textbf{B2-Blur}} & {\textbf{B2-JPEG}} & {\textbf{B2-Resize}} \\
\midrule
McCloskey18~\cite{McCloskey18} & {--} & 36.70 \pm 5.16 & \bfseries 59.36 \pm 28.81 & 13.49 \pm 0.55 & 2.09 \pm 1.00 & 13.33 \pm 5.70 & 2.88 \pm 1.75 & 11.97 \pm 5.89 \\
Nataraj19~\cite{Nataraj19}    & {--} & \bfseries 97.53 \pm 1.84 & 86.54 \pm 12.49 & 84.75 \pm 0.94 & 27.97 \pm 0.85 & 38.31 \pm 3.16 & 46.16 \pm 4.09 & 51.42 \pm 1.14 \\
Nowroozi22~\cite{Nowroozi22}  & {--} & \bfseries 84.94 \pm 1.32 & 67.87 \pm 1.75 & 84.55 \pm 0.97 & 22.47 \pm 1.69 & 29.00 \pm 1.26 & 32.43 \pm 1.66 & 42.39 \pm 2.87 \\
Song24-RGB~\cite{Song24}      & {--} & {--} & \bfseries 91.38 \pm 2.63 & 2.34 \pm 0.61 & 0.00 \pm 0.00 & 1.19 \pm 0.40 & 0.18 \pm 0.32 & 0.00 \pm 0.00 \\
\cmidrule{1-9}
Marra19a~\cite{Marra19a}      & {--} & \bfseries 100.00 \pm 0.00 & 40.71 \pm 5.20 & 9.55 \pm 2.10 & 1.67 \pm 1.07 & 13.50 \pm 9.26 & 4.10 \pm 1.06 & 5.04 \pm 1.18 \\
Durall20~\cite{Durall20}      & {--} & \bfseries 99.90 \pm 0.20 & 95.23 \pm 7.93 & 99.46 \pm 0.94 & 20.83 \pm 5.80 & 61.63 \pm 2.21 & 39.04 \pm 4.51 & 94.12 \pm 1.11 \\
Dzanic20~\cite{Dzanic20}      & {--} & 98.90 \pm 1.71 & 99.10 \pm 1.37 & \bfseries 99.78 \pm 0.22 & 40.33 \pm 0.18 & 90.59 \pm 4.08 & 48.26 \pm 0.54 & 98.63 \pm 0.61 \\
Giudice21~\cite{Giudice21}    & \bfseries 100.00 \pm 0.00 & {--} & {--} & 92.74 \pm 8.01 & 21.99 \pm 1.30 & 59.21 \pm 3.61 & 45.48 \pm 1.80 & 81.18 \pm 1.76 \\
Corvi23-R~\cite{Corvi23}      & \bfseries 100.00 \pm 0.00 & {--} & {--} & 35.16 \pm 3.84 & 2.44 \pm 1.03 & 17.02 \pm 4.26 & 1.11 \pm 0.37 & 25.20 \pm 6.53 \\
Corvi23-S~\cite{Corvi23}      & \bfseries 77.44 \pm 27.47 & {--} & {--} & 15.08 \pm 7.42 & 0.42 \pm 0.72 & 27.90 \pm 18.07 & 0.42 \pm 0.72 & 32.21 \pm 15.68 \\
Song24-Freq~\cite{Song24}     & {--} & {--} & 31.08 \pm 3.82 & \bfseries 70.56 \pm 6.94 & 5.30 \pm 0.87 & 50.55 \pm 3.78 & 6.79 \pm 1.55 & 68.64 \pm 0.63 \\
\cmidrule{1-9}
Qian20~\cite{Qian20}          & \bfseries 90.71 \pm 6.43 & {--} & {--} & 82.05 \pm 6.12 & 56.08 \pm 38.06 & 67.12 \pm 28.49 & 26.28 \pm 6.65 & 75.90 \pm 20.89 \\
Wang20~\cite{Wang20}          & \bfseries 100.00 \pm 0.00 & {--} & {--} & 78.19 \pm 18.90 & 0.67 \pm 0.77 & 37.08 \pm 54.53 & 34.17 \pm 57.01 & 49.04 \pm 44.23 \\
Song24-SL~\cite{Song24}       & {--} & {--} & \bfseries 84.45 \pm 0.85 & 55.86 \pm 1.97 & 2.68 \pm 0.74 & 20.18 \pm 1.66 & 7.77 \pm 1.51 & 34.61 \pm 2.59 \\
\bottomrule
\end{tabular}
}
\end{table*}

We evaluate all MFD methods against their compatible attack strategies, as summarized in Table~\ref{tab:fingerprint_method_properties}. ASRs are reported separately for removal (Table~\ref{tab:removal_attacks}) and forgery (Table~\ref{tab:forgery_attacks}) attacks, with the highest ASR per method shown in \textbf{bold}.

Removal attacks are broadly effective across most MFD methods. In many cases, the strongest attack achieves over $80\%$ ASR under the $\ell_\infty$ norm constraint, indicating that fingerprints can often be suppressed while maintaining high perceptual quality. Corresponding LPIPS and PSNR values remain within good image quality ranges for most attacks, and upon visual inspection of representative attacked images we confirm that the perturbations are imperceptible (see Tables~\ref{tab:removal_attacks_lpips}, \ref{tab:removal_attacks_psnr}, and Figure~\ref{fig:attack_visualization} in Appendix~\ref{apx:image_quality}).

The exceptions are Marra19a and Song24-RGB, which exhibit strong resistance to most attacks unless white-box access is available. McCloskey18, Corvi23-R, and Corvi23-S also show some resistance to black-box attacks; however, their baseline (non-adversarial) accuracies are substantially lower. Marra19a relies on noise residuals (image minus its BM3D-denoised version), and our results indicate that structured residuals are difficult to fully eliminate under black-box settings, making the method more robust. Song24-RGB measures the distance of a sample to the true data manifold in the RGB domain. Because fingerprints from the same model likely form tight clusters in this space and our attacks enforce an $\ell_\infty$ perturbation bound in the RGB domain, perturbed samples rarely escape their cluster, limiting the effectiveness of removal. In contrast, Song24-Freq and Song24-SL lack such constraints in the frequency or learned feature domains, which explains their greater vulnerability to black-box attacks compared to Song24-RGB. Since these methods are highly vulnerable to white-box attacks yet robust in black-box scenarios, it is likely that their fingerprints are inherently less transferable from the image space. Taken together, these findings suggest that noise residual- and manifold-based representations holds promise for robustness against removal in black-box scenarios.

Table~\ref{tab:removal_attacks} shows that ASR varies based on adversarial knowledge. White-box attacks (W1--W3) generally outperform black-box attacks (B1 and B2) across MFD techniques, which is expected since white-box attackers have privileged visibility into the MFD system. Among white-box attacks, W1 consistently achieves the best performance---when applicable. However, W1 attacks are not applicable for techniques that do not support direct gradient optimization. For these cases, W2 and W3 attacks rely on approximations of the gradients. As seen in the table, these approximations sometimes underperform relative to B1 strategies, which are also approximated. This counterintuitive result arises when the approximation errors of W2 and W3 exceed those of B1. 

Even when B1 is not the optimal strategy, its success remains competitive with white-box approaches for several MFD methods, showing that black-box surrogate-based attacks can achieve relatively high success rates with only knowledge of the MFD technique implementation.
In many applications, such as forensic use cases, MFD techniques cannot realistically be kept secret, making B1 removal attacks a practical threat.

Although adaptive strategies (W1--W3, and B1) always outperform fingerprint-agnostic strategies based on simple image manipulations (B2), these generic black-box perturbations alone often achieve ASRs $>50\%$, with one of the techniques (Dzanic20) being particularly vulnerable (ASR $>90\%$). This result indicates that fingerprint traces can be significantly disrupted without any knowledge of the underlying MFD technique, demonstrating the fragility of current MFD methods and raising concerns about the applicability of these techniques in adversarial scenarios.

ASRs show low standard deviation across most attack instances. However, some attacks exhibit high variance, with W3 against McCloskey18 showing up to $28.81$ pp.\ difference in ASR. The choice of MFD technique appears to be the main driver of this variance, creating uncertainty for users about attack robustness. This variability suggests certain samples are inherently more vulnerable to fingerprint removal. We leave a detailed investigation of the underlying causes for future work.

\begin{table}[!htbp]
\centering
\caption{Attack success rates (ASR, mean~$\pm$~std, in \%) of \textbf{forgery attacks}. White-box attacks include direct gradient-based optimization (W1), analytic approximations (W2), and surrogate fingerprint extractors (W3). Black-box attacks include surrogate classifiers (B1). The maximum value per method is highlighted in \textbf{bold}.}
\label{tab:forgery_attacks}
\resizebox{\columnwidth}{!}{
\setlength{\tabcolsep}{3pt}
\begin{tabular}{l 
    S[table-format=2.2(2.2)] 
    S[table-format=2.2(1.2)] 
    S[table-format=2.2(1.2)] 
    S[table-format=2.2(1.2)] 
}
\toprule
\textbf{Method} & {\textbf{W1}} & {\textbf{W2}} & {\textbf{W3}} & {\textbf{B1}} \\
\midrule
McCloskey18~\cite{McCloskey18} & {--} & 7.51 \pm 2.82 & \bfseries 10.43 \pm 2.81 & 2.10 \pm 0.06 \\
Nataraj19~\cite{Nataraj19}    & {--} & \bfseries 51.64 \pm 2.78 & 9.85 \pm 3.17 & 11.27 \pm 0.72 \\
Nowroozi22~\cite{Nowroozi22}  & {--} & \bfseries 12.09 \pm 3.57 & 6.84 \pm 0.73 & 10.20 \pm 1.57 \\
Song24-RGB~\cite{Song24}      & {--} & {--} & \bfseries 16.42 \pm 4.59 & 0.00 \pm 0.00 \\
\cmidrule{1-5}
Marra19a~\cite{Marra19a}      & {--} & \bfseries 97.85 \pm 2.90 & 4.46 \pm 1.28 & 0.50 \pm 0.52 \\
Durall20~\cite{Durall20}      & {--} & \bfseries 57.67 \pm 8.63 & 23.49 \pm 2.70 & 18.18 \pm 1.13 \\
Dzanic20~\cite{Dzanic20}      & {--} & 15.19 \pm 5.09 & \bfseries 25.53 \pm 9.61 & 25.51 \pm 3.93 \\
Giudice21~\cite{Giudice21}    & \bfseries 98.92 \pm 0.39 & {--} & {--} & 13.95 \pm 4.09 \\
Corvi23-R~\cite{Corvi23}      & \bfseries 99.77 \pm 0.40 & {--} & {--} & 3.67 \pm 1.38 \\
Corvi23-S~\cite{Corvi23}      & \bfseries 9.12 \pm 4.06 & {--} & {--} & 2.52 \pm 1.47 \\
Song24-Freq~\cite{Song24}     & {--} & {--} & 3.31 \pm 1.38 & \bfseries 6.66 \pm 0.88 \\
\cmidrule{1-5}
Qian20~\cite{Qian20}          & \bfseries 18.12 \pm 15.79 & {--} & {--} & 12.51 \pm 1.64 \\
Wang20~\cite{Wang20}          & \bfseries 99.55 \pm 0.77 & {--} & {--} & 5.68 \pm 5.09 \\
Song24-SL~\cite{Song24}       & {--} & {--} & \bfseries 10.09 \pm 1.90 & 7.29 \pm 1.47 \\
\bottomrule
\end{tabular}
}
\end{table}

\subsection{Robustness to Forgery}

Table~\ref{tab:forgery_attacks} shows that forgery ASRs are below those of removal across the board. Forgery is a fundamentally harder task than removal: with 12 models in the pool, successful forgery requires targeted misattribution to a specific model, while removal succeeds upon misclassification into \emph{any} incorrect class. Geometrically, removal corresponds to crossing the nearest decision boundary, whereas forgery requires pushing the image to a potentially farther target region.
When MFD considers only two classes, such as in deepfake detection, removal and forgery become equivalent: erasing traces of one source implies attribution to the other. But as the number of classes grows, the decision landscape becomes increasingly complex, making forgery attacks harder to execute.

Despite this difficulty, many methods still exhibit substantial forgery ASRs, especially in white-box settings. On average, W1 yields the highest ASRs, followed by W2 and then W3, consistent with their decreasing levels of gradient access—from direct gradients (W1) to analytic approximations (W2) to surrogate-based approximations (W3). Even in the black-box setting, ASRs reach up to 25.51\%, indicating that a non-trivial fraction of samples can still be intentionally misattributed to a chosen target model. Importantly, note that the reported means and standard deviations are computed by averaging over all 12 generative models. Refer to Figure~\ref{fig:model_vulnerability} for the variation in attack vulnerability of each model when used as a forgery target. In practice, the attacker may be interested in a specific model, or a specific model family that might be more vulnerable to attacks, in which case the average ASR within the family might be significantly higher. Furthermore, our gradient-based attacks are method-agnostic rather than exploiting the specific architecture of each detector; tailored attacks would likely achieve even higher forgery ASRs.

Similar to removal attacks, forged samples cannot be visually distinguished from their original counterparts, maintaining an LPIPS $<0.05$ and PSNR $>35$ dB across all forged samples (see Tables~\ref{tab:forgery_attacks_lpips}, \ref{tab:forgery_attacks_psnr} in Appendix~\ref{apx:image_quality}). It is important to note that even higher ASRs for both forgery and removal are feasible if the adversary relaxes the constraint on image quality ($\ell_{\infty}$ bound), as explained in Appendix~\ref{apx:hyperparameters}.

Recall that we discarded B2 attack strategies from the forgery attack evaluation because they lack specificity to embed a targeted model fingerprint into the forged sample. For the remaining attacks, we observe trends similar to those in the removal evaluation. As with removal, white-box settings tend to outperform B1 strategies, with W1 consistently achieving the highest ASRs. In contrast, ASR variance for McCloskey18 is relatively lower than in removal attacks, while Qian20 now exhibits the highest variance. This finding indicates that while the objectives of removal and forgery are similar, they are not perfectly complementary, and motivates a deeper investigation of the relationship between removal and forgery.

\subsection{Removal vs.\ Forgery}

\begin{figure}[!htbp]
    \centering
    \includegraphics[width=1.0\columnwidth]{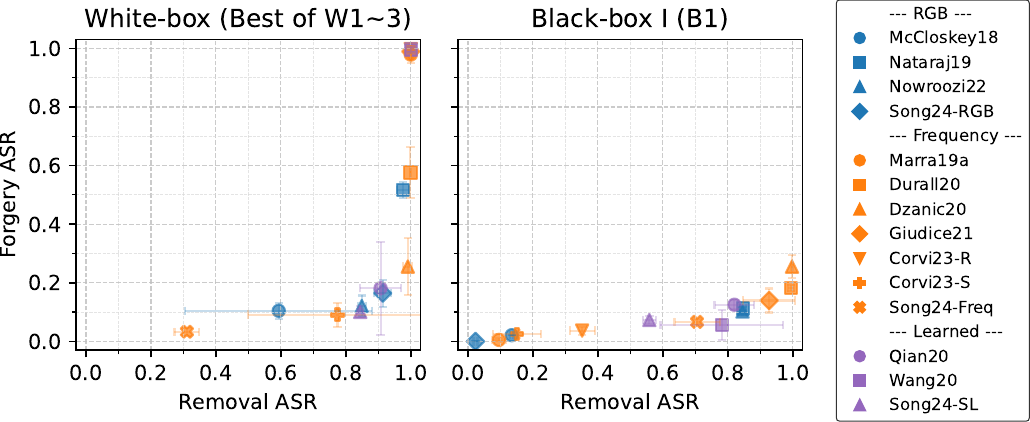}
    \caption{Correlation between removal and forgery attack vulnerabilities across MFD methods. Left: white-box scenario using the best ASR per method (highest ASR among W1, W2, and W3) for removal and forgery, capturing white-box vulnerability. Right: black-box I scenario (B1). Points are MFD methods; error bars show stdevs across runs. Forgery has no B2, so B2 is omitted. Colors encode fingerprint domain as shown in the legend box. See Figure~\ref{fig:corr_removal_vs_forgery} in Appendix~\ref{apx:removal_vs_forgery} for individual whit-box attack data.}
    \label{fig:corr_removal_vs_forgery_wbest_vs_b1}
\end{figure}

We observe a clear positive correlation between the ASR of removal and forgery, for both white-box and black-box scenarios. As shown in Fig.~\ref{fig:corr_removal_vs_forgery_wbest_vs_b1}, MFD methods that are more susceptible to removal are also generally more vulnerable to forgery. This trend is intuitive: since all reported attacks are gradient-based, if gradients can efficiently guide perturbations toward a distant forgery region, they can likewise push samples outside their current decision region, leading to removal.

\begin{figure*}[!htbp]
    \centering
    \includegraphics[width=0.9\textwidth]{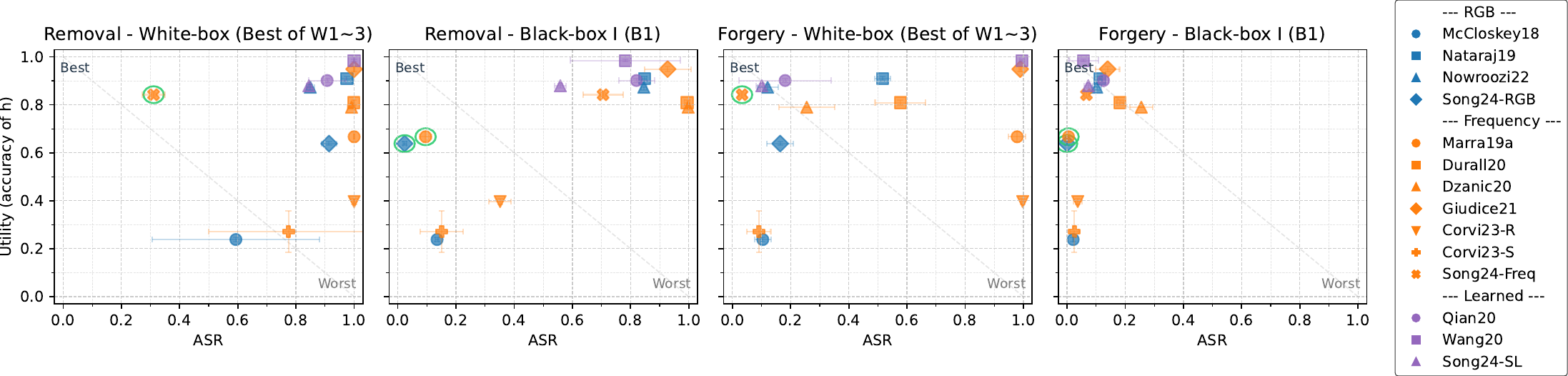}
    \caption{Trade-off between attribution utility (accuracy of the classifier $h$, y-axis) and robustness (ASR, x-axis) across four scenarios: removal under white-box (far left) and black-box I (second), and forgery under white-box (third) and black-box I (far right). White-box scenario uses the best ASR per method (highest ASR among W1, W2, and W3). Each point is an MFD method; error bars show standard deviation across runs. Methods nearer the top-left deliver stronger utility–robustness trade-offs. Green circles indicate outliers with better trade-off. See Figure~\ref{fig:tradeoff_utility_vs_asr_by_attack} in Appendix~\ref{apx:utility_robustness_trade_off} for a more detailed analysis on white-box attacks.}
    \label{fig:tradeoff_utility_vs_asr_wbest_vs_b1}
\end{figure*}

However, for the black-box scenario, this correlation does not imply that high removal ASR translates into high forgery ASR in absolute value terms. For B1 (Figure~\ref{fig:corr_removal_vs_forgery_wbest_vs_b1}, Right), we observed that the increase in forgery ASR from less-resilient to more-resilient methods remains small in magnitude. The magnitude of this increase likely depends on the size of the model pool and the characteristics of the models themselves. For instance, in the simplest case of a two-model pool, the attribution problem reduces to binary classification (similar to deepfake detection). In this setting, removing fingerprints of one model effectively forces the sample to be attributed to the other, making removal and forgery effectively the same goal. As the model pool grows, removal remains comparably feasible because pushing samples outside the true model's decision region does not depend on the other models. Forgery, by contrast, becomes increasingly difficult: it requires moving a sample into a specific target region, and reaching that region becomes harder with more decision boundaries.

Both removal and forgery ASR scales can be affected by the geometric landscape of decision regions across models in each individual model attribution task. The precise relationship between removal and forgery thus depends on the size of the model pool and the characteristics of the models. A more detailed analysis of how pool size and model diversity govern this relationship is left to closer examination.

\subsection{Utility--Robustness Trade-off}

We benchmark the original attribution accuracy (clean, non-adversarial) of all evaluated MFD methods, with results summarized in Table~\ref{tab:fingerprint_method_properties} under the ``Accuracy'' column. A few approaches, such as~\cite{Corvi23,McCloskey18}, were originally designed for binary deepfake detection (e.g., real vs.\ AI-generated images) rather than multi-class attribution. As a result, they offer limited utility for attribution tasks, despite the strong detection performance originally reported in the deepfake setting.

To examine the trade-off between attribution accuracy and adversarial robustness for MFD methods, we plot in Figure~\ref{fig:tradeoff_utility_vs_asr_wbest_vs_b1} this relationship for both removal and forgery attacks under white- and black-box scenarios. Broadly, we observe a utility--robustness tension for both removal and forgery attacks: methods with higher attribution accuracy tend to suffer higher ASR, indicating weaker robustness. Most high-utility methods (accuracy $>0.7$) exhibit significantly higher vulnerability than low-utility ones. 

For example, neural network–based methods such as Wang20 achieve the highest attribution accuracy but are also among the most vulnerable to both removal and forgery, particularly under white-box attacks. Nonetheless, there are notable exceptions: for instance, Marra19a and Song24-RGB under black-box attacks, and Song24-Freq under white-box attacks. These methods pair moderate attribution accuracy with substantially stronger robustness, making them stand out as the most balanced candidates among all the evaluated approaches.

Overall, these findings highlight a fundamental trade-off: methods with high attribution accuracy often lack robustness, whereas more robust methods typically sacrifice accuracy. The results indicate a few directions for future work, including: (i) improving residual- and manifold-based approaches such as Marra19a and Song24 to boost accuracy; (ii) strengthening CNN-based methods like Wang20 against adversarial perturbations; and (iii) exploring hybrid or ensemble designs that combine the robustness of residual/manifold methods with the strong attribution accuracy of CNN-based approaches.

\subsection{Attack Variance Across Models}

\begin{figure*}[!htbp]
    \centering
    \includegraphics[width=0.85\textwidth]{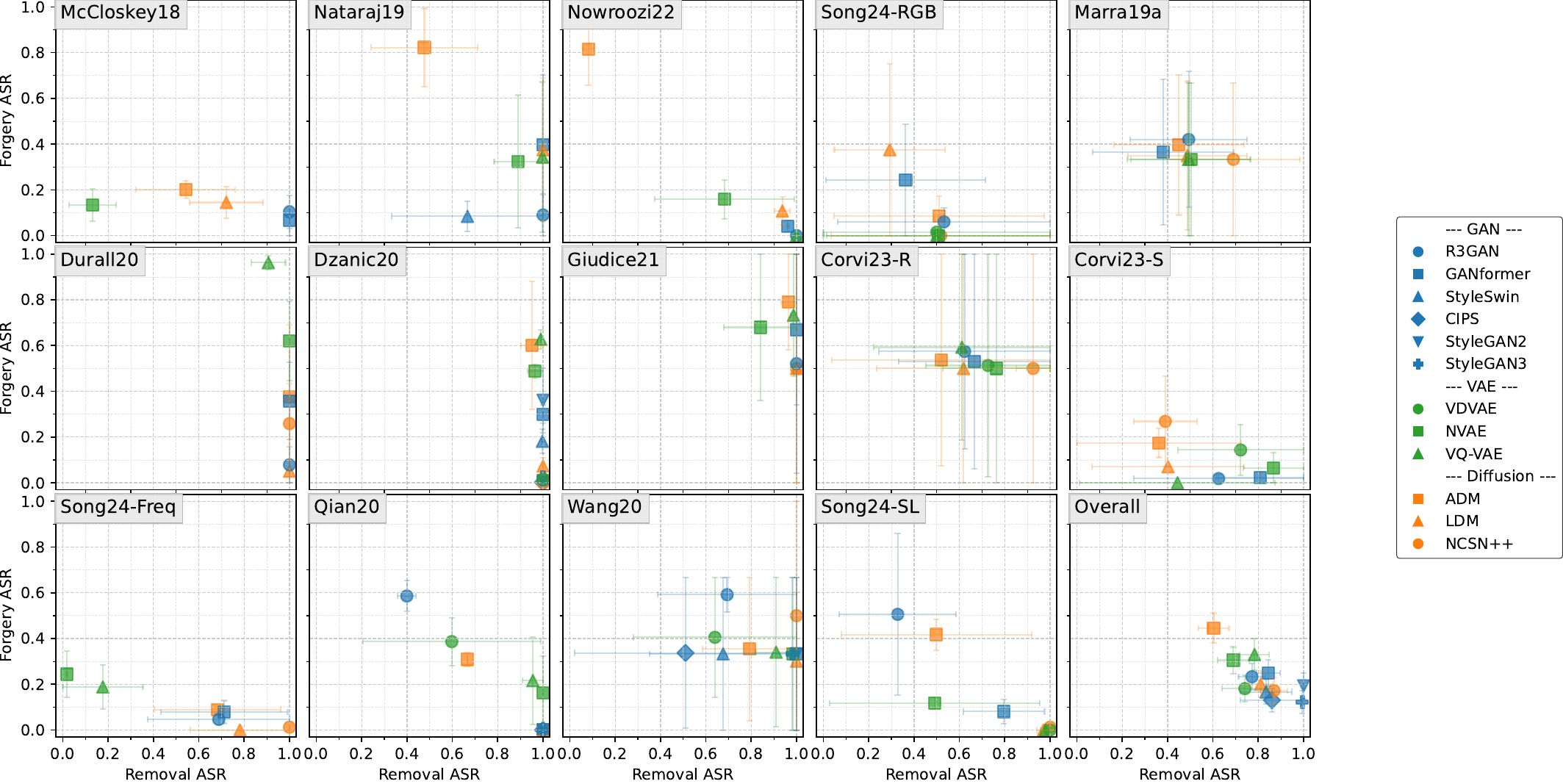}
    \caption{Model vulnerability per MFD method, averaged over all applicable attack types. Each subplot shows removal ASR ($x$) vs.\ forgery ASR ($y$) for the generative models (indicated by colors and markers) tied to that method; error bars show standard deviation across different attack types. The final subplot averages the data over all MFD methods (with standard deviation representing statistics over different MFD methods), showing a general vulnerability trend of generative models: models with more resilience to removal are more vulnerable to forgery, and vice versa.}
    \label{fig:model_vulnerability}
\end{figure*}

To examine whether particular generative models are inherently more vulnerable to certain adversarial attacks, we evaluated per-model vulnerability across all MFD methods. Figure \ref{fig:model_vulnerability} plots forgery ASR versus removal ASR for each model under each MFD method, averaging over all applicable attack strategies.

The results show that attack vulnerability is not consistently high or low for any given model or model family. Instead, it varies strongly with the MFD method. For example, in removal, NVAE is among the least vulnerable under McCloskey18 and Song24-Freq, yet appears highly vulnerable under other methods. Likewise, ADM is markedly more robust in Nowroozi22 than all other models, but not in any of the remaining methods. In forgery, ADM is significantly more vulnerable under Nataraj19 and Nowroozi22, yet this pattern does not generalize across all MFD methods.

Despite this variability, a trend emerges: across MFD methods, image generation models that are more robust to removal tend to be more vulnerable to forgery, and vice versa. This trade-off is visible in the final subplot of Figure \ref{fig:model_vulnerability}, and is also particularly clear under methods such as Nowroozi22, Song24-Freq, and Qian20. Intuitively, if a model’s fingerprint occupies a more isolated decision region, it may be easy for gradients to escape that region (removal) but harder to steer perturbations into it (forgery). We further observe that diffusion and VAE models are more resistant to removal and more vulnerable to forgery, whereas GAN-based models show the opposite trend. We leave a deeper investigation of these behaviors to future work.

In summary, vulnerability patterns in attribution attacks cannot be attributed solely to the generative model architecture nor solely to the MFD method used for attribution. Instead, they arise from the interaction between the two, jointly shaping the attribution decision landscape.

\section{Discussion}
\label{sec:discussion}

This paper asks a practical question: \emph{how well does model fingerprint detection (MFD) hold up when the content creator actively tries to evade attribution?}
We address this by evaluating 14 MFD methods on a 12-model attribution task under three attacker knowledge levels (White-box, Black-box Access I, and Black-box Access II), two attacker goals (removal and forgery), and five attack strategies.

Our results support three main conclusions.
(1) \emph{Removal attacks are effective.} For most MFD methods, the attacker can remove the fingerprints with imperceptible perturbations, even without information about the model or the MFD technique (Table~II).
(2) \emph{Forgery is relatively less effective but should not be dismissed.} We still observe significant forgery success for many MFD methods, especially in white-box settings (Table~III).
(3) \emph{High clean accuracy does not imply robustness.} Our results uncover a tension between clean attribution performance and adversarial robustness: none of the evaluated MFD methods is both consistently accurate and robust across the threat models we consider.

\subsection{Implications for Provenance Use-cases}
Many MFD papers motivate attribution as a provenance mechanism.
Our findings suggest an important caveat: \emph{a method that performs well on clean benchmarks is unlikely to achieve the same performance in adversarial settings.}
This is especially relevant in forensic applications, where the subjects of an investigation have incentives to remove or forge fingerprints.
Based on our evaluation, existing MFD methods are not yet ready for deployment in these contexts.
A practical takeaway is that the evaluation methodology of MFD techniques in such applications should consider robustness, and should clearly define the threat model considered.

It is also important to evaluate attribution under more realistic candidate sets. Real-world deployments would face large, open-world scenarios and frequent model updates that introduce closely related variants, such as fine-tuned checkpoints. Studying how attacks scale to realistic candidate pools is essential for understanding the practical effectiveness of MFD techniques, both with and without adversarial interference.

\subsection{What the Robustness Patterns Suggest}
The patterns in Tables~II--III indicate that robustness is largely method-driven.
Some MFD approaches appear relatively more resistant in constrained black-box settings, especially residual- and manifold-based MFD methods. However, this robustness is not universal and may not hold under white-box access.
A useful way to view this is that black-box robustness can reflect limited \emph{attack transferability} under a particular perturbation budget, rather than a fundamental inability to manipulate the fingerprint.
This distinction matters operationally: if the attribution pipeline is public, it has leaked, or can be closely approximated, the threat of adaptive attacks becomes more relevant.

These observations suggest a clear design objective for future research in MFD: methods should be developed and reported with an explicit robustness target, rather than focusing solely on clean accuracy. Based on the utility-robustness trade-off perspective, we identify two promising directions: (i) improving the clean accuracy of inherently robust representations, potentially leveraging NN-based classifiers to boost performance; and (ii) hardening high-accuracy fingerprints---such as MFD methods based on NN-learned features---against adaptive perturbations without sacrificing clean performance, potentially through techniques like adversarial training or randomized smoothing~\cite{cohen2019certified}.

\subsection{Limitations and Future Work}
First, we examine multi-class \emph{attribution}, a task inherently more complex than binary deepfake detection. A critical distinction lies in the relationship between attack types: while untargeted removal and targeted forgery are functionally equivalent in binary detection, they diverge significantly in the attribution setting. In multi-class attribution, untargeted removal is generally easier to achieve than evading detection of a deepfake because the adversary can exploit any of the multiple decision boundaries surrounding the correct class manifold. In contrast, targeted forgery remains more difficult, as it requires a precise shift into a specific, potentially distant region of the feature space. Consequently, an attack that successfully breaks attribution (removal) may still leave the sample within the broader manifold of synthetic content, failing to evade deepfake detection. Future work could study the gap between ``detection-robustness'' and ``attribution-robustness'' under the same threat model.

Second, while our attacks are strong and cover multiple access levels, they do not span the full space of adaptive strategies.
Future work could explore broader perturbation families and multi-step post-processing pipelines, and should also study whether fingerprint suppression or forgery leaves detectable traces that could be used as a warning signal that a verifier can flag for a more careful inspection.

Third, the scope of our evaluation is constrained by the specific generator and dataset configuration (12 generators trained on FFHQ at $256 \times 256$ resolution). 
Evaluating a broader range of generators and datasets, higher resolutions, and diverse types of distribution shifts would help quantify robustness under more diverse deployment conditions and thus provide a more complete understanding of MFD robustness.

Finally, our results suggest that provenance systems should not rely on a single mechanism alone.
Watermarking can provide a strong and unambiguous provenance signal when a watermark is embedded at generation time and remains detectable, and in such cases it is natural to rely primarily on the watermark.
However, watermarking is often unavailable in practice: many images are generated by open-source or legacy models without watermarking, detection may require secret keys or vendor-side access, and real-world distribution or editing pipelines can remove or break watermark detectability altogether.
Passive fingerprinting addresses a different need.
It does not require cooperation at generation time and is detected post hoc, which allows provenance analysis even when watermarking is unavailable or not possible.
Fingerprinting is therefore not a universal replacement for watermarking, nor is watermarking a replacement for fingerprinting.
It is thus worthwhile exploring hybrid provenance systems that use watermarking as a high-confidence signal when present, and falls back to fingerprinting when it is not, while explicitly accounting for removal, forgery, and uncertainty under adversarial manipulation~\cite{yang2024gaussian,gunn2024undetectable}.

\section{Conclusion}
\label{sec:conclusion}

Model fingerprint detection has been proposed as a scalable means of providing provenance and accountability for AI-generated images, with many methods reporting strong attribution accuracy in non-adversarial evaluations.
In this paper, we provide the first systematic security evaluation of such techniques for \emph{model attribution}, formalizing three adversary knowledge levels and benchmarking 14 representative methods against five attack strategies under two attacker objectives: fingerprint removal and fingerprint forgery.

Our results reveal a critical gap between clean and adversarial performance.
Fingerprint removal is highly effective for most methods through small, imperceptible perturbations, even in black-box regimes. Forgery, in contrast, is less effective but can still succeed in black-box settings and achieves high success under white-box access on some methods.
We also identify a utility--robustness trade-off: techniques with the highest clean attribution accuracy are frequently among the most vulnerable, and no evaluated method achieves both high accuracy and high robustness across all threat models.

These findings suggest that current passive fingerprinting approaches alone are insufficient for high-stakes provenance in adversarial settings.
Advancing reliable attribution will likely require (i) more robust fingerprint representations (e.g., residual/manifold-based), (ii) incorporating security considerations in the development and evaluation of new techniques, and (iii) hybrid designs that combine complementary provenance mechanisms, including active approaches like watermarking.

\section*{Usage of LLMs}
LLMs were used for editorial purposes in this manuscript, and all outputs were inspected by the authors to ensure accuracy and originality.

\section*{Acknowledgments}
We thank our anonymous reviewers for their valuable feedback, which substantially improved the paper.
This work was supported by the Edinburgh International Data Facility (EIDF) and the Data-Driven Innovation Programme at the University of Edinburgh. Access to EIDF was facilitated through the University of Edinburgh's Generative AI Laboratory GAIL Fellow scheme.
Marc Juarez is a GAIL fellow and the recipient of a Google Research Scholar Award in Security.

\bibliographystyle{IEEEtran}
\bibliography{reference.bib}

@article{McCloskey18,
  author    = {Scott McCloskey and Michael Albright},
  title     = {Detecting {GAN}-Generated Imagery using Color Cues},
  journal   = {arXiv preprint arXiv:1812.08247},
  year      = {2018}
}

@inproceedings{Nataraj19,
  author    = {Lakshmanan Nataraj and Tajuddin Manhar Mohammed and Shivkumar Chandrasekaran and Arjuna Flenner and Jawadul~H. Bappy and Amit~K. Roy-Chowdhury and B.~S. Manjunath},
  title     = {Detecting {GAN} Generated Fake Images using Co-Occurrence Matrices},
  booktitle = {IS\&T Electronic Imaging, Media Watermarking, Security, and Forensics 2019},
  year      = {2019},
  pages     = {532-1--532-7},
  doi       = {10.2352/ISSN.2470-1173.2019.5.MWSF-532}
}

@inproceedings{Durall20,
  author    = {Ricard Durall and Margret Keuper and Janis Keuper},
  title     = {Watch Your Up-Convolution: {CNN} Based Generative Deep Neural Networks Are Failing to Reproduce Spectral Distributions},
  booktitle = {Proceedings of the IEEE/CVF Conference on Computer Vision and Pattern Recognition (CVPR)},
  year      = {2020},
  pages     = {7890--7899},
  doi       = {10.1109/CVPR42600.2020.00791}
}

@inproceedings{Dzanic20,
  author    = {Tarik Dzanic and Karan Shah and Freddie Witherden},
  title     = {Fourier Spectrum Discrepancies in Deep Network Generated Images},
  booktitle = {Advances in Neural Information Processing Systems, Vol. 33 (NeurIPS)},
  year      = {2020},
  pages     = {3022--3032}
}

@inproceedings{Wang20,
  author    = {Sheng{-}Yu Wang and Oliver Wang and Richard Zhang and Andrew Owens and Alexei~A. Efros},
  title     = {{CNN}-Generated Images Are Surprisingly Easy to Spot...for Now},
  booktitle = {Proceedings of the IEEE/CVF Conference on Computer Vision and Pattern Recognition (CVPR)},
  year      = {2020},
  pages     = {8692--8701},
  doi       = {10.1109/CVPR42600.2020.00872}
}

@inproceedings{Marra18,
  author    = {Francesco Marra and Diego Gragnaniello and Davide Cozzolino and Luisa Verdoliva},
  title     = {Detection of {GAN}-Generated Fake Images over Social Networks},
  booktitle = {2018 IEEE Conference on Multimedia Information Processing and Retrieval (MIPR)},
  year      = {2018},
  pages     = {384--389}
}

@inproceedings{Marra19b,
  author    = {Francesco Marra and Cristiano Saltori and Giulia Boato and Luisa Verdoliva},
  title     = {Incremental Learning for the Detection and Classification of {GAN}-Generated Images},
  booktitle = {2019 IEEE International Workshop on Information Forensics and Security (WIFS)},
  year      = {2019},
  pages     = {1--6}
}

@inproceedings{Yu19,
  author    = {Ning Yu and Larry~S. Davis and Mario Fritz},
  title     = {Attributing Fake Images to {GAN}s: Learning and Analyzing {GAN} Fingerprints},
  booktitle = {Proceedings of the IEEE/CVF International Conference on Computer Vision (ICCV)},
  year      = {2019},
  pages     = {7555--7565},
  doi       = {10.1109/ICCV.2019.00765}
}

@inproceedings{Song24,
  author    = {Hae Jin Song and Mahyar Khayatkhoei and Wael AbdAlmageed},
  title     = {{ManiFPT}: Defining and Analyzing Fingerprints of Generative Models},
  booktitle = {Proceedings of the IEEE/CVF Conference on Computer Vision and Pattern Recognition (CVPR)},
  year      = {2024},
  note      = {to appear}
}

@inproceedings{Qian20,
  title        = {{Thinking in Frequency: Face Forgery Detection by Mining Frequency-aware Clues}},
  author       = {Qian, Yuyang and Yin, Guojun and Sheng, Lu and Chen, Zixuan and Shao, Jing},
  booktitle    = {European Conference on Computer Vision (ECCV) Workshops},
  year         = {2020},
}

@article{Giudice21,
  title        = {{Fighting deepfakes by detecting GAN DCT anomalies}},
  author       = {Giudice, Oliver and Guarnera, Luca and Battiato, Sebastiano},
  journal      = {arXiv preprint},
  year         = {2021},
}

@incollection{Nowroozi22,
  title={Detecting high-quality GAN-generated face images using neural networks},
  author={Nowroozi, Ehsan and Mekdad, Yassine},
  booktitle={Big Data Analytics and Intelligent Systems for Cyber Threat Intelligence},
  pages={235--252},
  year={2023},
  publisher={River Publishers}
}

@inproceedings{Corvi23,
  title        = {{Intriguing properties of synthetic images: From generative adversarial networks to diffusion models}},
  author       = {Corvi, Riccardo and Cozzolino, Davide and Poggi, Giovanni and Nagano, Koki and Verdoliva, Luisa},
  booktitle    = {CVPR Workshops (CVPRW)},
  year         = {2023},
}

@INPROCEEDINGS{Marra19a,
  author={Marra, Francesco and Gragnaniello, Diego and Verdoliva, Luisa and Poggi, Giovanni},
  booktitle={2019 IEEE Conference on Multimedia Information Processing and Retrieval (MIPR)}, 
  title={Do GANs Leave Artificial Fingerprints?}, 
  year={2019},
  volume={},
  number={},
  pages={506-511},
  doi={10.1109/MIPR.2019.00103}}

@inproceedings{Guarnera20,
  title={Deepfake detection by analyzing convolutional traces},
  author={Guarnera, Luca and Giudice, Oliver and Battiato, Sebastiano},
  booktitle={Proceedings of the IEEE/CVF conference on computer vision and pattern recognition workshops},
  pages={666--667},
  year={2020}
}

@inproceedings{Girish21,
  title={Towards discovery and attribution of open-world gan generated images},
  author={Girish, Sharath and Suri, Saksham and Rambhatla, Sai Saketh and Shrivastava, Abhinav},
  booktitle={Proceedings of the IEEE/CVF international conference on computer vision},
  pages={14094--14103},
  year={2021}
}

@inproceedings{Frank20,
  title={Leveraging frequency analysis for deep fake image recognition},
  author={Frank, Joel and Eisenhofer, Thorsten and Sch{\"o}nherr, Lea and Fischer, Asja and Kolossa, Dorothea and Holz, Thorsten},
  booktitle={International conference on machine learning},
  pages={3247--3258},
  year={2020},
  organization={PMLR}
}

@article{Song25,
  title={Riemannian-Geometric Fingerprints of Generative Models},
  author={Song, Hae Jin and Itti, Laurent},
  journal={arXiv preprint arXiv:2506.22802},
  year={2025}
}

@article{Goebel20,
  title={Adversarial attacks on co-occurrence features for GAN detection},
  author={Goebel, Michael and Manjunath, BS},
  journal={arXiv preprint arXiv:2009.07456},
  year={2020}
}

@inproceedings{Wesselkamp22,
  title={Misleading deep-fake detection with gan fingerprints},
  author={Wesselkamp, Vera and Rieck, Konrad and Arp, Daniel and Quiring, Erwin},
  booktitle={2022 IEEE Security and Privacy Workshops (SPW)},
  pages={59--65},
  year={2022},
  organization={IEEE}
}

@inproceedings{Karras2020ADA,
  author = {Karras, Tero and Aittala, Miika and Hellsten, Janne and Laine, Samuli and Lehtinen, Jaakko and Aila, Timo},
  title = {Training Generative Adversarial Networks with Limited Data},
  booktitle = {Advances in Neural Information Processing Systems (NeurIPS)},
  year = {2020}
}

@inproceedings{Karras2021StyleGAN3,
  author = {Karras, Tero and Aittala, Miika and Laine, Samuli and Härkönen, Erik and Hellsten, Janne and Lehtinen, Jaakko and Aila, Timo},
  title = {Alias‑Free Generative Adversarial Networks},
  booktitle = {Advances in Neural Information Processing Systems (NeurIPS)},
  year = {2021}
}

@inproceedings{Hudson2021GANformer,
  author = {Hudson, Drew A. and others},
  title = {Generative Adversarial Transformers},
  booktitle = {International Conference on Machine Learning (ICML)},
  year = {2021},
  volume = {139}
}

@inproceedings{Zhang2022StyleSwin,
  author = {Zhang, Bowen and others},
  title = {StyleSwin: Transformer‑Based GAN for High‑Resolution Image Generation},
  booktitle = {IEEE/CVF Conference on Computer Vision and Pattern Recognition (CVPR)},
  year = {2022}
}

@inproceedings{Huang2024R3GAN,
  author = {Huang, Yiwen and Gokaslan, Aaron and Kuleshov, Volodymyr and Tompkin, James},
  title = {The GAN Is Dead; Long Live the GAN! A Modern GAN Baseline},
  booktitle = {Advances in Neural Information Processing Systems (NeurIPS)},
  year = {2024}
}

@inproceedings{Anokhin2021CIPS,
  author = {Anokhin, Ilya and others},
  title = {Image Generators With Conditionally‑Independent Pixel Synthesis},
  booktitle = {IEEE/CVF Conference on Computer Vision and Pattern Recognition (CVPR)},
  year = {2021},
  note = {Oral Presentation}
}

@inproceedings{Child2021VDVAE,
  author = {Child, Rewon},
  title = {Very Deep VAEs Generalize Autoregressive Models and Can Outperform Them on Images},
  booktitle = {International Conference on Learning Representations (ICLR)},
  year = {2021}
}

@inproceedings{Song2021ScoreSDE,
  author = {Song, Yang and Sohl‑Dickstein, Jascha and Kingma, Diederik P. and Kumar, Abhishek and Ermon, Stefano and Poole, Ben},
  title = {Score‑Based Generative Modeling Through Stochastic Differential Equations},
  booktitle = {International Conference on Learning Representations (ICLR)},
  year = {2021}
}

@article{an2024waves,
  title={Waves: Benchmarking the robustness of image watermarks},
  author={An, Bang and Ding, Mucong and Rabbani, Tahseen and Agrawal, Aakriti and Xu, Yuancheng and Deng, Chenghao and Zhu, Sicheng and Mohamed, Abdirisak and Wen, Yuxin and Goldstein, Tom and others},
  journal={arXiv preprint arXiv:2401.08573},
  year={2024}
}

@article{lin2025crack,
  title={A Crack in the Bark: Leveraging Public Knowledge to Remove Tree-Ring Watermarks},
  author={Lin, Junhua and Juarez, Marc},
  journal={arXiv preprint arXiv:2506.10502},
  year={2025}
}

@article{goebel2020adversarial,
  author    = {Michael Goebel and B.~S. Manjunath},
  title     = {Adversarial Attacks on Co-Occurrence Features for {GAN} Detection},
  journal   = {arXiv preprint arXiv:2009.07456},
  year      = {2020}
}

@article{Hu2024StableSignature,
  title   = {Stable Signature is Unstable: Removing Image Watermark from Diffusion Models},
  author  = {Yuepeng Hu and Zhengyuan Jiang and Moyang Guo and Neil Z. Gong},
  journal = {arXiv preprint arXiv:2405.07145},
  year    = {2024}
}

@inproceedings{Jiang2023EvadingWatermark,
  title     = {Evading Watermark Based Detection of AI-Generated Content},
  author    = {Zhengyuan Jiang and Jinghuai Zhang and Neil Z. Gong},
  booktitle = {Proc. ACM Conf. on Computer and Communications Security (CCS)},
  year      = {2023}
}

@inproceedings{Zhao2024InvisibleRemovable,
  title     = {Invisible Image Watermarks Are Provably Removable Using Generative AI},
  author    = {Xuandong Zhao and Kexun Zhang and Zihao Su and Saastha Vasan and Ilya Grishchenko and Christopher Kruegel and Giovanni Vigna and Yu-Xiang Wang and Lei Li},
  booktitle = {Advances in Neural Information Processing Systems (NeurIPS)},
  year      = {2024}
}

@inproceedings{bird2023typology,
  title={Typology of risks of generative text-to-image models},
  author={Bird, Charlotte and Ungless, Eddie and Kasirzadeh, Atoosa},
  booktitle={Proceedings of the 2023 AAAI/ACM Conference on AI, Ethics, and Society},
  pages={396--410},
  year={2023}
}

@article{bbc_deepnude_2019,
  author    = {BBC News},
  title     = {App that generated fake naked images shut down},
  year      = {2019},
  url       = {https://www.bbc.co.uk/news/technology-48799045},
  note      = {Accessed: Sep 18, 2025},
}

@misc{adobe2023firefly,
  title        = {Adobe Firefly: Generative AI for Creative Tools},
  author       = {{Adobe Inc.}},
  year         = {2023},
  howpublished = {\url{https://business.adobe.com/ai/adobe-genai.html}},
  note         = {Accessed: 2025-09-19}
}

@article{katsamakas2024gcp,
  title   = {Generative Artificial Intelligence, Content Creation, and Platforms},
  author  = {Katsamakas, Emmanuel and others},
  journal = {Journal of International Economics \& Trade},
  year    = {2024},
  volume  = {24},
  number  = {1},
  pages   = {1--20},
  doi     = {10.1007/s10842-024-00430-9}
}

@misc{openai2023dalle3,
  title        = {{DALL{\textperiodcentered}E 3 API Documentation}},
  author       = {{OpenAI}},
  year         = {2023},
  howpublished = {\url{https://help.openai.com/en/articles/8555480-dall-e-3-api}},
  note         = {Accessed: 2025-09-19}
}

@article{goodfellow2014explaining,
  title={Explaining and harnessing adversarial examples},
  author={Goodfellow, Ian J and Shlens, Jonathon and Szegedy, Christian},
  journal={arXiv preprint arXiv:1412.6572},
  year={2014}
}

@inproceedings{zhang2018unreasonable,
  title={The unreasonable effectiveness of deep features as a perceptual metric},
  author={Zhang, Richard and Isola, Phillip and Efros, Alexei A and Shechtman, Eli and Wang, Oliver},
  booktitle={Proceedings of the IEEE conference on computer vision and pattern recognition},
  pages={586--595},
  year={2018}
}

@article{madry2017towards,
  title={Towards deep learning models resistant to adversarial attacks},
  author={Madry, Aleksander and Makelov, Aleksandar and Schmidt, Ludwig and Tsipras, Dimitris and Vladu, Adrian},
  journal={arXiv preprint arXiv:1706.06083},
  year={2017}
}

@inproceedings{Levy2014NeuralWordEmbedding,
  title     = {Neural Word Embedding as Implicit Matrix Factorization},
  author    = {Levy, Omer and Goldberg, Yoav},
  booktitle = {Advances in Neural Information Processing Systems},
  volume    = {27},
  year      = {2014}
}

@article{Yusuf2020DifferentiableHistogram,
  title   = {Differentiable Histogram with Hard-Binning},
  author  = {Yusuf, Ismail and others},
  journal = {arXiv preprint arXiv:2012.06311},
  year    = {2020}
}

@inproceedings{zhang2018lpips,
  title={The Unreasonable Effectiveness of Deep Features as a Perceptual Metric},
  author={Zhang, Richard and Isola, Phillip and Efros, Alexei A. and Shechtman, Eli and Wang, Oliver},
  booktitle={Proceedings of the IEEE Conference on Computer Vision and Pattern Recognition (CVPR)},
  pages={586--595},
  year={2018}
}

@article{huynh2008psnr,
  title={Scope of Validity of {PSNR} in Image/Video Quality Assessment},
  author={Huynh-Thu, Quan and Ghanbari, Mohammed},
  journal={Electronics Letters},
  volume={44},
  number={13},
  pages={800--801},
  year={2008},
  publisher={IET}
}

@article{boenisch2021systematic,
  title={A systematic review on model watermarking for neural networks},
  author={Boenisch, Franziska},
  journal={Frontiers in big Data},
  volume={4},
  pages={729663},
  year={2021},
  publisher={Frontiers Media SA}
}

@inproceedings{yang2024gaussian,
  title={Gaussian shading: Provable performance-lossless image watermarking for diffusion models},
  author={Yang, Zijin and Zeng, Kai and Chen, Kejiang and Fang, Han and Zhang, Weiming and Yu, Nenghai},
  booktitle={Proceedings of the IEEE/CVF Conference on Computer Vision and Pattern Recognition},
  pages={12162--12171},
  year={2024}
}

@article{gunn2024undetectable,
  title={An undetectable watermark for generative image models},
  author={Gunn, Sam and Zhao, Xuandong and Song, Dawn},
  journal={arXiv preprint arXiv:2410.07369},
  year={2024}
}

@article{van2017neural,
  title={Neural discrete representation learning},
  author={Van Den Oord, Aaron and Vinyals, Oriol and others},
  journal={Advances in neural information processing systems},
  volume={30},
  year={2017}
}

@inproceedings{rombach2022high,
  title={High-resolution image synthesis with latent diffusion models},
  author={Rombach, Robin and Blattmann, Andreas and Lorenz, Dominik and Esser, Patrick and Ommer, Bj{\"o}rn},
  booktitle={Proceedings of the IEEE/CVF conference on computer vision and pattern recognition},
  pages={10684--10695},
  year={2022}
}

@article{dhariwal2021diffusion,
  title={Diffusion models beat gans on image synthesis},
  author={Dhariwal, Prafulla and Nichol, Alexander},
  journal={Advances in neural information processing systems},
  volume={34},
  pages={8780--8794},
  year={2021}
}

@article{vahdat2020nvae,
  title={NVAE: A deep hierarchical variational autoencoder},
  author={Vahdat, Arash and Kautz, Jan},
  journal={Advances in neural information processing systems},
  volume={33},
  pages={19667--19679},
  year={2020}
}

@inproceedings{cohen2019certified,
  title={Certified adversarial robustness via randomized smoothing},
  author={Cohen, Jeremy and Rosenfeld, Elan and Kolter, Zico},
  booktitle={international conference on machine learning},
  pages={1310--1320},
  year={2019},
  organization={PMLR}
}

\newpage
\appendices
\section{Image Quality Under Attacks}
\label{apx:image_quality}

\begin{table*}[!htbp]
\centering
\caption{Perceptual distances (LPIPS, mean~$\pm$~std) of \textbf{removal attacks} against evaluated MFD methods. 
White-box attacks include direct gradient-based optimization (W1), analytic approximations (W2), and surrogate fingerprint extractors (W3). 
Black-box attacks include surrogate classifiers (B1) and fingerprint-agnostic perturbations (B2: Gaussian noise, blur, JPEG compression, resizing). 
Lower LPIPS indicates better perceptual quality preservation.}
\label{tab:removal_attacks_lpips}
\resizebox{\textwidth}{!}{
\begin{tabular}{lcccccccc}
\toprule
\textbf{Method} & \textbf{W1} & \textbf{W2} & \textbf{W3} & \textbf{B1} & \textbf{B2-Noise} & \textbf{B2-Blur} & \textbf{B2-JPEG} & \textbf{B2-Resize} \\
\midrule
McCloskey18~\cite{McCloskey18} & -- & 0.0261 $\pm$ 0.0011 & 0.0247 $\pm$ 0.0016 & 0.0270 $\pm$ 0.0012 & 0.0246 $\pm$ 0.0007 & 0.0176 $\pm$ 0.0006 & 0.0286 $\pm$ 0.0006 & 0.0175 $\pm$ 0.0004 \\
Nataraj19~\cite{Nataraj19}    & -- & 0.0238 $\pm$ 0.0010 & 0.0242 $\pm$ 0.0013 & 0.0263 $\pm$ 0.0011 & 0.0147 $\pm$ 0.0003 & 0.0247 $\pm$ 0.0005 & 0.0320 $\pm$ 0.0006 & 0.0262 $\pm$ 0.0002 \\
Nowroozi22~\cite{Nowroozi22}  & -- & 0.0251 $\pm$ 0.0019 & 0.0246 $\pm$ 0.0015 & 0.0285 $\pm$ 0.0020 & 0.0153 $\pm$ 0.0003 & 0.0242 $\pm$ 0.0005 & 0.0312 $\pm$ 0.0005 & 0.0252 $\pm$ 0.0005 \\
Song24-RGB~\cite{Song24}   & -- & -- & 0.0239 $\pm$ 0.0005 & 0.0275 $\pm$ 0.0016 & 0.0146 $\pm$ 0.0003 & 0.0248 $\pm$ 0.0004 & 0.0320 $\pm$ 0.0005 & 0.0263 $\pm$ 0.0007 \\
\cmidrule{1-9}
Marra19a~\cite{Marra19a}    & -- & 0.0256 $\pm$ 0.0010 & 0.0263 $\pm$ 0.0014 & 0.0271 $\pm$ 0.0009 & 0.0147 $\pm$ 0.0004 & 0.0249 $\pm$ 0.0004 & 0.0321 $\pm$ 0.0002 & 0.0260 $\pm$ 0.0003 \\
Durall20~\cite{Durall20}    & -- & 0.0268 $\pm$ 0.0020 & 0.0275 $\pm$ 0.0018 & 0.0259 $\pm$ 0.0012 & 0.0143 $\pm$ 0.0005 & 0.0255 $\pm$ 0.0005 & 0.0322 $\pm$ 0.0007 & 0.0271 $\pm$ 0.0003 \\
Dzanic20~\cite{Dzanic20}    & -- & 0.0227 $\pm$ 0.0011 & 0.0269 $\pm$ 0.0014 & 0.0254 $\pm$ 0.0010 & 0.0153 $\pm$ 0.0004 & 0.0246 $\pm$ 0.0005 & 0.0316 $\pm$ 0.0004 & 0.0253 $\pm$ 0.0003 \\
Giudice21~\cite{Giudice21}   & 0.0272 $\pm$ 0.0010 & -- & -- & 0.0265 $\pm$ 0.0008 & 0.0147 $\pm$ 0.0004 & 0.0248 $\pm$ 0.0006 & 0.0322 $\pm$ 0.0004 & 0.0263 $\pm$ 0.0004 \\
Corvi23-R~\cite{Corvi23}    & 0.0281 $\pm$ 0.0009 & -- & -- & 0.0258 $\pm$ 0.0007 & 0.0168 $\pm$ 0.0003 & 0.0243 $\pm$ 0.0006 & 0.0323 $\pm$ 0.0005 & 0.0250 $\pm$ 0.0004 \\
Corvi23-S~\cite{Corvi23}    & 0.0265 $\pm$ 0.0025 & -- & -- & 0.0249 $\pm$ 0.0020 & 0.0263 $\pm$ 0.0005 & 0.0206 $\pm$ 0.0008 & 0.0354 $\pm$ 0.0004 & 0.0220 $\pm$ 0.0004 \\
Song24-Freq~\cite{Song24}  & -- & -- & 0.0236 $\pm$ 0.0011 & 0.0267 $\pm$ 0.0006 & 0.0146 $\pm$ 0.0002 & 0.0247 $\pm$ 0.0005 & 0.0319 $\pm$ 0.0006 & 0.0257 $\pm$ 0.0007 \\
\cmidrule{1-9}
Qian20~\cite{Qian20}      & 0.0284 $\pm$ 0.0003 & -- & -- & 0.0269 $\pm$ 0.0010 & 0.0141 $\pm$ 0.0001 & 0.0254 $\pm$ 0.0002 & 0.0313 $\pm$ 0.0001 & 0.0260 $\pm$ 0.0003 \\
Wang20~\cite{Wang20}      & 0.0278 $\pm$ 0.0002 & -- & -- & 0.0272 $\pm$ 0.0068 & 0.0142 $\pm$ 0.0003 & 0.0254 $\pm$ 0.0001 & 0.0310 $\pm$ 0.0005 & 0.0264 $\pm$ 0.0002 \\
Song24-SL~\cite{Song24}    & -- & -- & 0.0241 $\pm$ 0.0050 & 0.0274 $\pm$ 0.0043 & 0.0145 $\pm$ 0.0004 & 0.0248 $\pm$ 0.0003 & 0.0316 $\pm$ 0.0002 & 0.0257 $\pm$ 0.0003 \\
\bottomrule
\end{tabular}
}
\end{table*}

\begin{table*}[!htbp]
\centering
\caption{Peak signal-to-noise ratio (PSNR, mean~$\pm$~std, in dB) of \textbf{removal attacks} against evaluated MFD methods. 
White-box attacks include direct gradient-based optimization (W1), analytic approximations (W2), and surrogate fingerprint extractors (W3). 
Black-box attacks include surrogate classifiers (B1) and fingerprint-agnostic perturbations (B2: Gaussian noise, blur, JPEG compression, resizing). 
Higher PSNR indicates better perceptual quality preservation.}
\label{tab:removal_attacks_psnr}
\resizebox{\textwidth}{!}{
\begin{tabular}{lcccccccc}
\toprule
\textbf{Method} & \textbf{W1} & \textbf{W2} & \textbf{W3} & \textbf{B1} & \textbf{B2-Noise} & \textbf{B2-Blur} & \textbf{B2-JPEG} & \textbf{B2-Resize} \\
\midrule
McCloskey18~\cite{McCloskey18} & -- & 37.4012 $\pm$ 0.0674 & 36.2441 $\pm$ 0.5015 & 36.7965 $\pm$ 0.0203 & 42.0924 $\pm$ 0.0251 & 40.2988 $\pm$ 0.1299 & 37.5891 $\pm$ 0.0749 & 40.0573 $\pm$ 0.1292 \\
Nataraj19~\cite{Nataraj19}    & -- & 35.6142 $\pm$ 0.0021 & 36.9385 $\pm$ 0.1912 & 35.9843 $\pm$ 0.0593 & 41.8021 $\pm$ 0.0061 & 40.5872 $\pm$ 0.0541 & 37.1992 $\pm$ 0.0330 & 40.1264 $\pm$ 0.0562 \\
Nowroozi22~\cite{Nowroozi22}  & -- & 35.8034 $\pm$ 0.8610 & 36.9541 $\pm$ 0.2201 & 35.9012 $\pm$ 0.6812 & 41.8362 $\pm$ 0.0171 & 40.6510 $\pm$ 0.1518 & 37.2914 $\pm$ 0.0903 & 40.2135 $\pm$ 0.1486 \\
Song24-RGB~\cite{Song24}   & -- & -- & 37.0224 $\pm$ 0.0049 & 36.8610 $\pm$ 2.0319 & 41.8074 $\pm$ 0.0091 & 40.5244 $\pm$ 0.0443 & 37.1814 $\pm$ 0.0201 & 40.0691 $\pm$ 0.0369 \\
\cmidrule{1-9}
Marra19a~\cite{Marra19a}    & -- & 35.1841 $\pm$ 0.0571 & 35.5762 $\pm$ 0.6992 & 35.6635 $\pm$ 0.0078 & 41.7992 $\pm$ 0.0061 & 40.5863 $\pm$ 0.0694 & 37.2075 $\pm$ 0.0294 & 40.1234 $\pm$ 0.0619 \\
Durall20~\cite{Durall20}    & -- & 35.2911 $\pm$ 0.6684 & 35.4324 $\pm$ 0.6834 & 35.4681 $\pm$ 0.2861 & 41.7613 $\pm$ 0.0198 & 40.6843 $\pm$ 0.0889 & 37.2412 $\pm$ 0.0626 & 40.2044 $\pm$ 0.0953 \\
Dzanic20~\cite{Dzanic20}    & -- & 35.3134 $\pm$ 1.7921 & 35.2542 $\pm$ 0.0094 & 36.8073 $\pm$ 0.0165 & 41.8053 $\pm$ 0.0095 & 40.6964 $\pm$ 0.0778 & 37.3095 $\pm$ 0.0386 & 40.2531 $\pm$ 0.0710 \\
Giudice21~\cite{Giudice21}   & 35.2413 $\pm$ 0.0061 & -- & -- & 35.2544 $\pm$ 0.0096 & 41.7913 $\pm$ 0.0063 & 40.5821 $\pm$ 0.0561 & 37.1963 $\pm$ 0.0310 & 40.1192 $\pm$ 0.0549 \\
Corvi23-R~\cite{Corvi23}    & 35.2704 $\pm$ 0.0171 & -- & -- & 35.5209 $\pm$ 0.0579 & 41.8264 $\pm$ 0.0187 & 40.6084 $\pm$ 0.0891 & 37.2962 $\pm$ 0.0478 & 40.1794 $\pm$ 0.0809 \\
Corvi23-S~\cite{Corvi23}    & 35.3019 $\pm$ 0.0304 & -- & -- & 35.4150 $\pm$ 0.0094 & 41.8374 $\pm$ 0.0224 & 41.8022 $\pm$ 0.2835 & 38.1401 $\pm$ 0.1474 & 41.4723 $\pm$ 0.2899 \\
Song24-Freq~\cite{Song24}  & -- & -- & 36.9454 $\pm$ 0.1692 & 35.4734 $\pm$ 0.0063 & 41.8013 $\pm$ 0.0060 & 40.5971 $\pm$ 0.0561 & 37.2188 $\pm$ 0.0317 & 40.1374 $\pm$ 0.0548 \\
\cmidrule{1-9}
Qian20~\cite{Qian20}      & 35.2674 $\pm$ 0.0012 & -- & -- & 35.4744 $\pm$ 0.4161 & 41.8034 $\pm$ 0.0060 & 40.6004 $\pm$ 0.0012 & 37.1884 $\pm$ 0.0011 & 40.1184 $\pm$ 0.0009 \\
Wang20~\cite{Wang20}      & 35.2685 $\pm$ 0.0013 & -- & -- & 35.5591 $\pm$ 0.4302 & 41.8103 $\pm$ 0.0011 & 40.6022 $\pm$ 0.0009 & 37.1871 $\pm$ 0.0008 & 40.1181 $\pm$ 0.0008 \\
Song24-SL~\cite{Song24}    & -- & -- & 35.8811 $\pm$ 0.8534 & 35.4968 $\pm$ 0.1459 & 41.8101 $\pm$ 0.0090 & 40.5662 $\pm$ 0.0604 & 37.2110 $\pm$ 0.0313 & 40.1133 $\pm$ 0.0580 \\
\bottomrule
\end{tabular}
}
\end{table*}

\begin{table}[!htbp]
\centering
\caption{Perceptual distances (LPIPS, mean~$\pm$~std) of \textbf{forgery attacks} against evaluated MFD methods. 
White-box attacks include direct gradient-based optimization (W1), analytic approximations (W2), and surrogate fingerprint extractors (W3). 
Black-box attacks include surrogate classifiers (B1). 
Lower LPIPS indicates better perceptual quality preservation.}
\label{tab:forgery_attacks_lpips}
\resizebox{\columnwidth}{!}{
\begin{tabular}{lcccc}
\toprule
\textbf{Method} & \textbf{W1} & \textbf{W2} & \textbf{W3} & \textbf{B1} \\
\midrule
McCloskey18~\cite{McCloskey18} & -- & 0.0264 $\pm$ 0.0011 & 0.0258 $\pm$ 0.0010 & 0.0271 $\pm$ 0.0012 \\
Nataraj19~\cite{Nataraj19}    & -- & 0.0243 $\pm$ 0.0003 & 0.0238 $\pm$ 0.0003 & 0.0246 $\pm$ 0.0003 \\
Nowroozi22~\cite{Nowroozi22}  & -- & 0.0252 $\pm$ 0.0012 & 0.0247 $\pm$ 0.0013 & 0.0256 $\pm$ 0.0013 \\
Song24-RGB~\cite{Song24}   & -- & -- & 0.0241 $\pm$ 0.0007 & 0.0228 $\pm$ 0.0008 \\
\cmidrule{1-5}
Marra19a~\cite{Marra19a}    & -- & 0.0257 $\pm$ 0.0009 & 0.0249 $\pm$ 0.0008 & 0.0244 $\pm$ 0.0006 \\
Durall20~\cite{Durall20}    & -- & 0.0246 $\pm$ 0.0030 & 0.0239 $\pm$ 0.0018 & 0.0232 $\pm$ 0.0029 \\
Dzanic20~\cite{Dzanic20}    & -- & 0.0250 $\pm$ 0.0010 & 0.0263 $\pm$ 0.0012 & 0.0257 $\pm$ 0.0011 \\
Giudice21~\cite{Giudice21}   & 0.0268 $\pm$ 0.0009 & -- & -- & 0.0246 $\pm$ 0.0007 \\
Corvi23-R~\cite{Corvi23}    & 0.0284 $\pm$ 0.0012 & -- & -- & 0.0261 $\pm$ 0.0014 \\
Corvi23-S~\cite{Corvi23}    & 0.0269 $\pm$ 0.0020 & -- & -- & 0.0279 $\pm$ 0.0009 \\
Song24-Freq~\cite{Song24}  & -- & -- & 0.0240 $\pm$ 0.0007 & 0.0243 $\pm$ 0.0009 \\
\cmidrule{1-5}
Qian20~\cite{Qian20}      & 0.0265 $\pm$ 0.0002 & -- & -- & 0.0244 $\pm$ 0.0002 \\
Wang20~\cite{Wang20}      & 0.0271 $\pm$ 0.0002 & -- & -- & 0.0240 $\pm$ 0.0002 \\
Song24-SL~\cite{Song24}    & -- & -- & 0.0251 $\pm$ 0.0010 & 0.0264 $\pm$ 0.0102 \\
\bottomrule
\end{tabular}
}
\end{table}

\begin{table}[!htbp]
\centering
\caption{Peak signal-to-noise ratio (PSNR, mean~$\pm$~std, in dB) of \textbf{forgery attacks} against evaluated MFD methods. 
White-box attacks include direct gradient-based optimization (W1), analytic approximations (W2), and surrogate fingerprint extractors (W3). 
Black-box attacks include surrogate classifiers (B1). 
Higher PSNR indicates better perceptual quality preservation.}
\label{tab:forgery_attacks_psnr}
\resizebox{\columnwidth}{!}{
\begin{tabular}{lcccc}
\toprule
\textbf{Method} & \textbf{W1} & \textbf{W2} & \textbf{W3} & \textbf{B1} \\
\midrule
McCloskey18~\cite{McCloskey18} & -- & 37.1634 $\pm$ 0.0439 & 37.0112 $\pm$ 0.0265 & 36.9873 $\pm$ 0.0224 \\
Nataraj19~\cite{Nataraj19}    & -- & 36.7484 $\pm$ 0.0011 & 37.0125 $\pm$ 0.0093 & 36.7191 $\pm$ 0.0050 \\
Nowroozi22~\cite{Nowroozi22}  & -- & 36.8803 $\pm$ 0.0268 & 37.0182 $\pm$ 0.0204 & 36.7652 $\pm$ 0.0156 \\
Song24-RGB~\cite{Song24}   & -- & -- & 37.0231 $\pm$ 0.0058 & 37.2444 $\pm$ 0.0342 \\
\cmidrule{1-5}
Marra19a~\cite{Marra19a}    & -- & 36.7693 $\pm$ 0.0199 & 36.7671 $\pm$ 0.0187 & 36.7175 $\pm$ 0.0122 \\
Durall20~\cite{Durall20}    & -- & 36.8368 $\pm$ 0.0539 & 36.7484 $\pm$ 0.0273 & 36.7133 $\pm$ 0.0548 \\
Dzanic20~\cite{Dzanic20}    & -- & 36.3375 $\pm$ 0.0248 & 36.7914 $\pm$ 0.0200 & 36.7445 $\pm$ 0.0075 \\
Giudice21~\cite{Giudice21}   & 36.8524 $\pm$ 0.0089 & -- & -- & 36.7091 $\pm$ 0.0090 \\
Corvi23-R~\cite{Corvi23}    & 36.7491 $\pm$ 0.0090 & -- & -- & 36.7804 $\pm$ 0.0092 \\
Corvi23-S~\cite{Corvi23}    & 35.9128 $\pm$ 0.0638 & -- & -- & 36.7703 $\pm$ 0.0141 \\
Song24-Freq~\cite{Song24}  & -- & -- & 37.0042 $\pm$ 0.0075 & 36.7284 $\pm$ 0.0089 \\
\cmidrule{1-5}
Qian20~\cite{Qian20}      & 36.7428 $\pm$ 0.0048 & -- & -- & 36.6938 $\pm$ 0.0090 \\
Wang20~\cite{Wang20}      & 36.7263 $\pm$ 0.0061 & -- & -- & 36.7602 $\pm$ 0.0051 \\
Song24-SL~\cite{Song24}    & -- & -- & 36.7611 $\pm$ 0.0659 & 36.7183 $\pm$ 0.1011 \\
\bottomrule
\end{tabular}
}
\end{table}

\para{LPIPS and PSNR indicators.} To complement the evaluation of attack performance, we also assess the visual quality of adversarially perturbed images. We report LPIPS, a perceptual similarity metric, and PSNR, a pixel-level fidelity measure, as two widely adopted indicators of image quality under perturbations. Tables~\ref{tab:removal_attacks_lpips} and~\ref{tab:removal_attacks_psnr} summarize the results for removal attacks, while Tables~\ref{tab:forgery_attacks_lpips} and~\ref{tab:forgery_attacks_psnr} present the corresponding results for forgery attacks. Across both attack goals and all attack types and MFD methods, LPIPS values remain below 0.05 and PSNR values consistently exceed 35~dB, indicating that adversarial perturbations preserve high image fidelity.

\begin{sidewaysfigure*}
    \centering
    \includegraphics[width=\textheight]{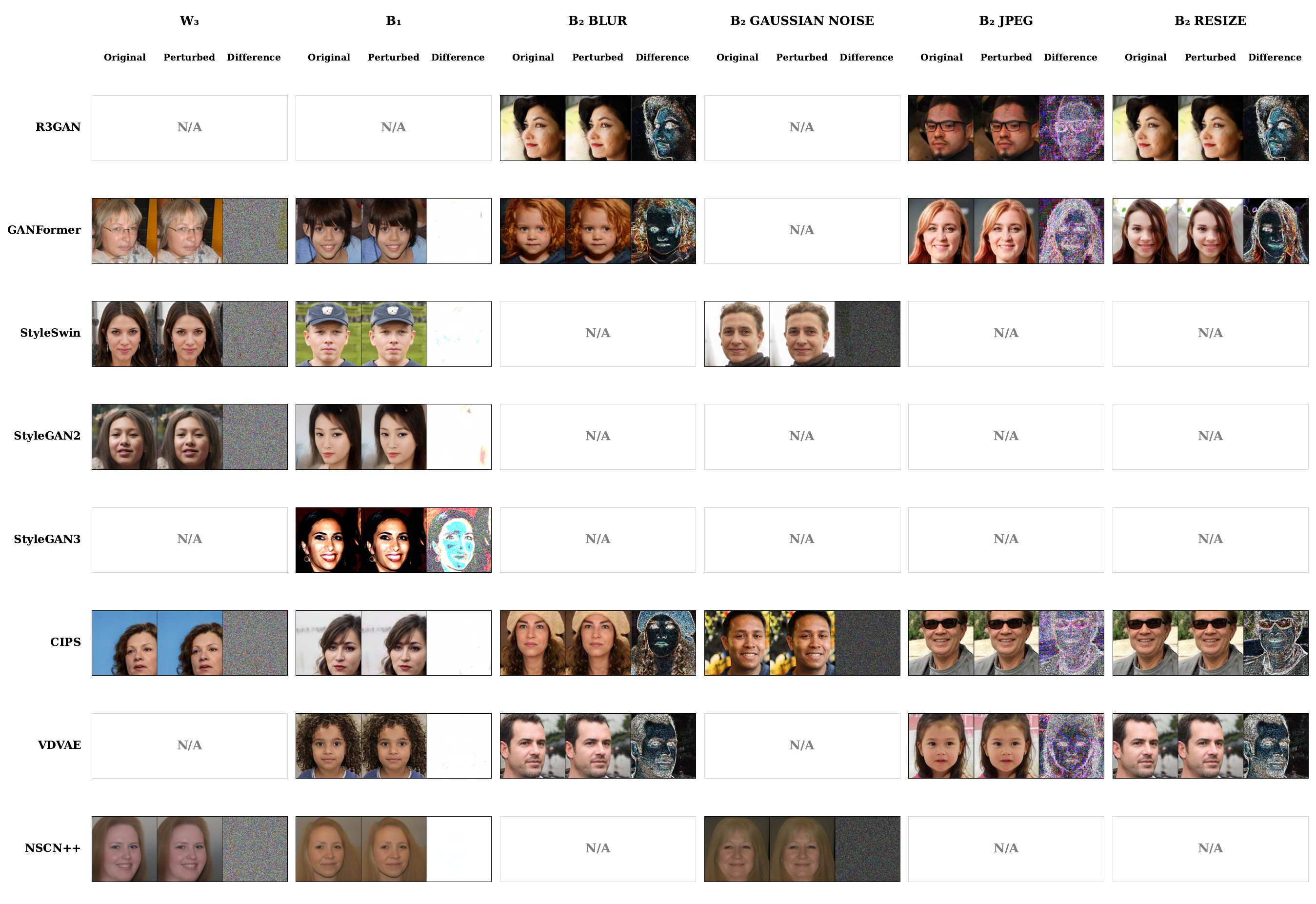}
    \caption{Visualization of successful removal attacks against the Song24-Freq MFD method for selected models. Each row represents a different generative model (target for fingerprint removal), and each column represents a different attack type. Within each cell, the three images show (from left to right): the original image, the perturbed image after attack, and the scaled difference. The difference is linearly mapped such that the lowest value corresponds to black and the highest value corresponds to white, with intermediate values scaled accordingly. The attack types include W3 (surrogate extractor), B1 (surrogate classifier), and B2 (basic image perturbations: blur, Gaussian noise, JPEG compression, and resize). The visualization demonstrates the effectiveness of various attack strategies in removing model fingerprints while maintaining visual quality. An ``N/A'' value indicates that there is no successfully attacked image for that entry.}
    \label{fig:attack_visualization}
\end{sidewaysfigure*}

\para{Example Images.} As an example, Figure~\ref{fig:attack_visualization} provides a comprehensive visualization of successful removal attacks against the Song24-Freq MFD method across different generative models and attack strategies. The grid structure clearly illustrates how various attack types (W3, B1, and B2 with different perturbation techniques) perform on different target models, with each cell showing the original image, the perturbed result, and the difference, which has been scaled to improve visibility. This visualization demonstrates the effectiveness of different attack strategies in removing model fingerprints while preserving the visual quality of the generated images. The difference images show the spatial distribution of perturbations introduced by each attack.

\newpage

\section{Removal ASR vs Forgery ASR}
\label{apx:removal_vs_forgery}

\begin{figure*}[!htbp]
    \centering
    \includegraphics[width=1.0\textwidth]{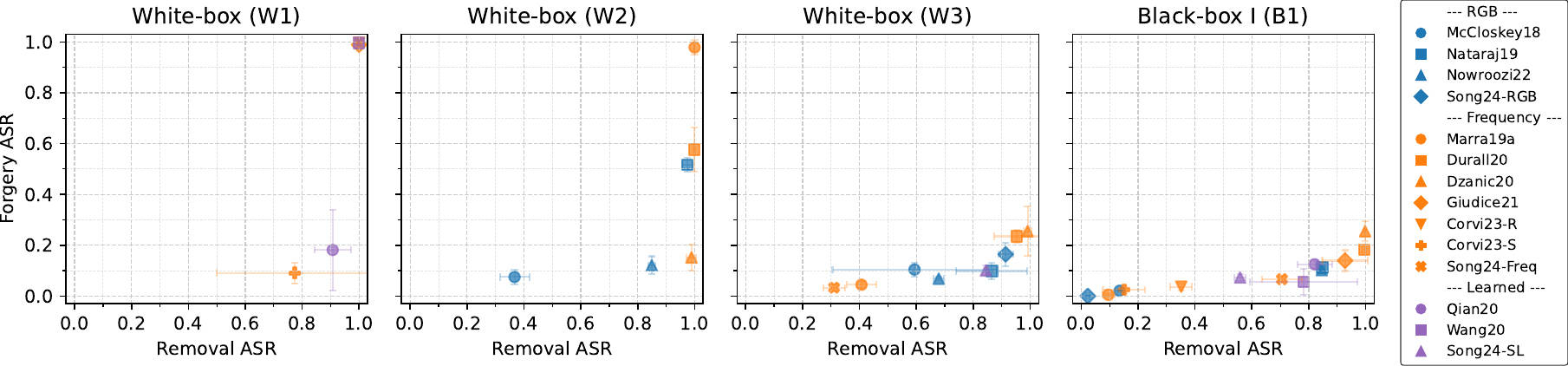}
    \caption{Correlation between removal and forgery attack vulnerabilities across MFD methods. We plot the data with regard to each attack type separately in this figure, as opposed to Figure~\ref{fig:corr_removal_vs_forgery_wbest_vs_b1}. Points are MFD methods; error bars show standard deviation across 5 independent runs. Forgery has no B2, so B2 is omitted. Colors encode fingerprint domain: blue=RGB, orange=frequency, purple=NN.}
    \label{fig:corr_removal_vs_forgery}
\end{figure*}

In Figure~\ref{fig:corr_removal_vs_forgery} we report the relationship between removal ASR and forgery ASR for each gradient-based attack type separately, rather than aggregating them using the best white-box ASR as done in the main text (Figure~\ref{fig:corr_removal_vs_forgery_wbest_vs_b1}). Specifically, we present results for W1, W2, and W3 individually. Note that certain attack types are only applicable to specific fingerprinting methods (e.g., W1 requires differentiability and is therefore only compatible with MFD methods such as Giudice21, Corvi23-R, Corvi23-S, Qian20, and Wang20). Importantly, the strong positive correlation between removal and forgery ASR persists for each independent attack type, reinforcing the trend reported in the main text.

\section{Utility-Robustness Trade-off}
\label{apx:utility_robustness_trade_off}

\begin{figure*}[!htbp]
    \centering
    \includegraphics[width=1.0\textwidth]{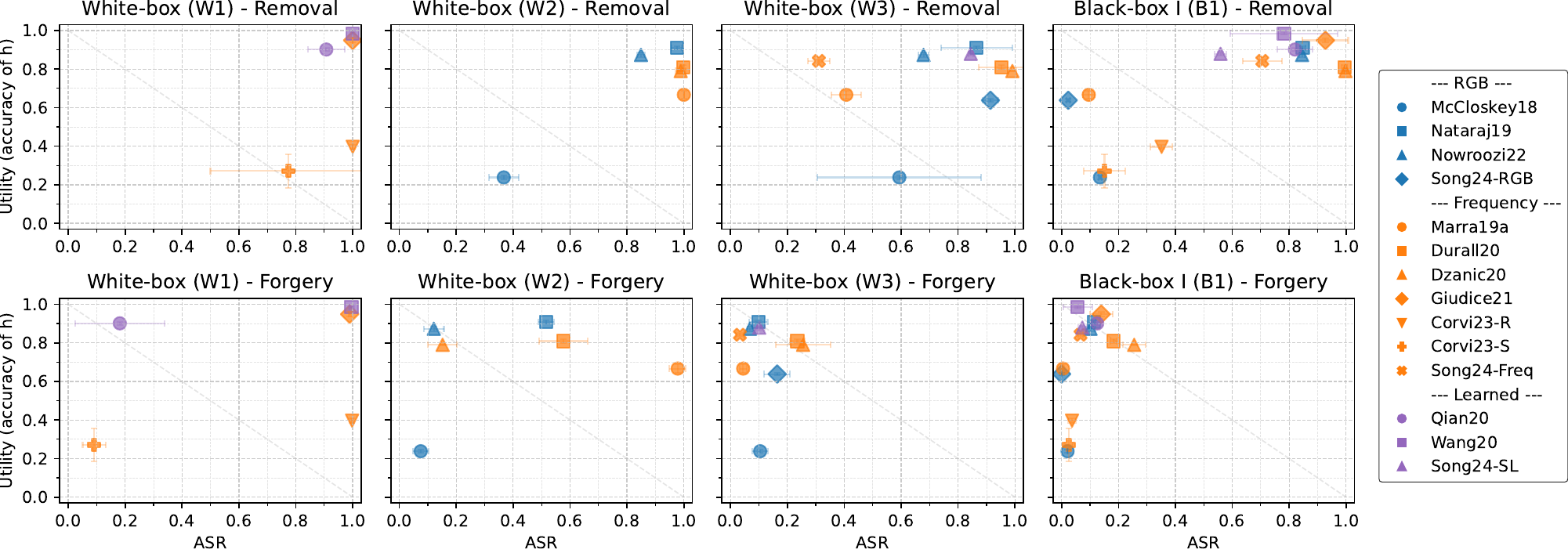}
    \caption{Trade-off between attribution utility (accuracy of classifier $h$, $y$-axis) and robustness (ASR, $x$-axis). We plot the data with regard to each attack type separately, as opposed to Figure~\ref{fig:tradeoff_utility_vs_asr_wbest_vs_b1}. Each point is an MFD method; error bars show standard deviation across 5 independent runs. Methods nearer the top-left achieve a better compromise between utility and robustness.}
    \label{fig:tradeoff_utility_vs_asr_by_attack}
\end{figure*}

In Figure~\ref{fig:tradeoff_utility_vs_asr_by_attack} we report the trade-off between utility and attack ASR for each gradient-based attack type separately, rather than aggregating them using the best white-box ASR as done in the main text (Figure~\ref{fig:tradeoff_utility_vs_asr_wbest_vs_b1}). Specifically, we present results for W1, W2, and W3 individually. Note that certain attack types are only applicable to specific fingerprinting methods.

\section{Hyperparameter Tuning}
\label{apx:hyperparameters}

\begin{figure*}[!htbp]
    \centering
    \includegraphics[width=1.0\textwidth]{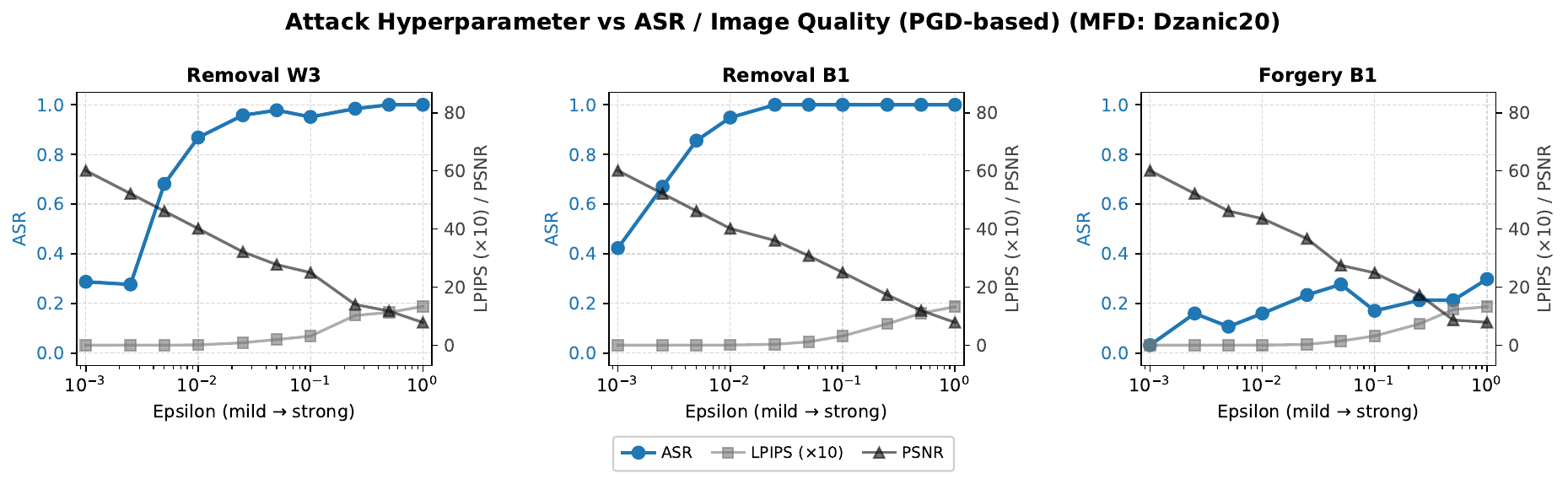}
    \caption{Effect of perturbation strength (PGD's $\epsilon$) on attack success and image quality for removal (W3, B1) and forgery (B1) attacks. The curves show that there is a trade-off between robustness and image quality regulated by perturbation strength.}
    \label{fig:ablation_pgd_panel}
\end{figure*}

\begin{figure*}[!htbp]
    \centering
    \includegraphics[width=1.0\textwidth]{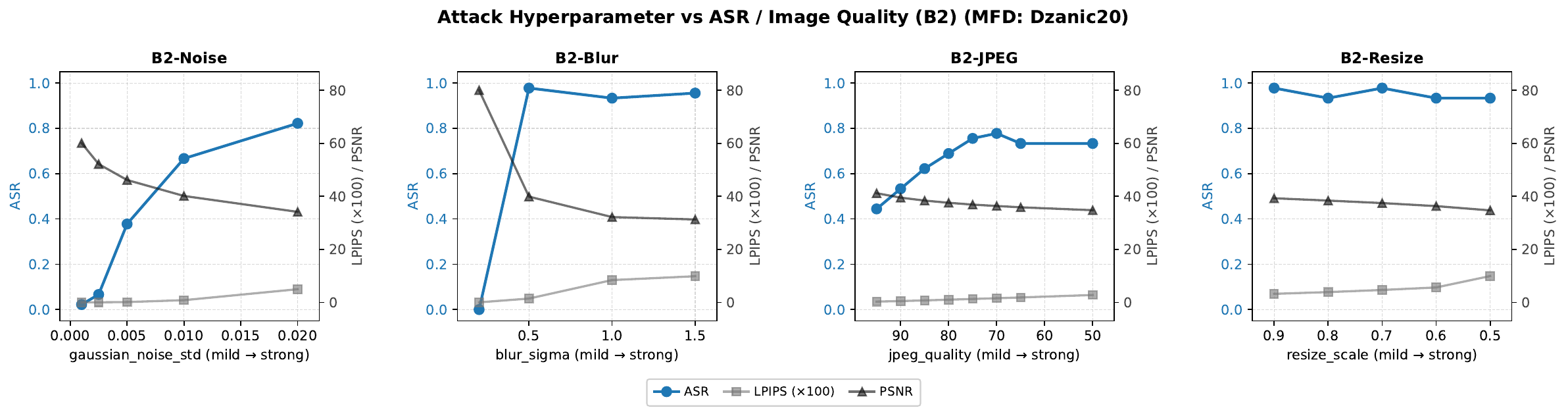}
    \caption{Attack hyperparameter study for B2 image perturbations (hyperparameters: JPEG quality, resize scale, blur $\sigma$, Gaussian noise std). X-axis: perturbation strength (mild$\to$strong). The curves illustrate the trade-off between attack success and image quality.}
    \label{fig:ablation_panel}
\end{figure*}

We conducted hyperparameter tuning for several attack cases with respect to their strength parameters to illustrate the trade-off between ASR and image quality degradation. Specifically, we report ablation results for multiple attack cases using Dzanic20 as a representative example, as shown in Figures~\ref{fig:ablation_pgd_panel} and~\ref{fig:ablation_panel}.

For PGD-based attacks (W3 and B1), we varied the perturbation budget $\epsilon$ (the maximum allowed change per pixel) over the range \{0.001, 0.005, 0.01, 0.05, 0.1, 0.5, 1.0\}. 

For image-perturbation (B2) attacks, we swept each perturbation parameter from mild to strong settings:
\begin{itemize}
    \item Gaussian noise standard deviation: \{0.001, 0.0025, 0.005, 0.01, 0.02\};
    \item Gaussian blur sigma: \{0.2, 0.5, 1.0, 1.5\};
    \item JPEG quality: \{95, 90, 85, 80, 75, 70, 65, 50\};
    \item Resize scaling factor: \{0.9, 0.8, 0.7, 0.6, 0.5\}.
\end{itemize}

Figures~\ref{fig:ablation_pgd_panel} and~\ref{fig:ablation_panel} plot ASR and image quality indicators (LPIPS/PSNR) as functions of these hyperparameters. As expected, increasing attack strength generally yields higher ASR, but at the cost of noticeably degraded image quality, particularly for PGD-based attacks.

\end{document}